\documentclass[10pt]{article}
\usepackage[utf8]{inputenc}
\usepackage[T1]{fontenc}
\usepackage{amsmath,amssymb,amsfonts}
\usepackage{graphicx}
\usepackage[margin=1in]{geometry}
\usepackage{hyperref}
\usepackage{float}
\usepackage{caption}
\usepackage{booktabs}
\usepackage{tabularx}
\usepackage{abstract}
\usepackage{titlesec}
\usepackage{enumitem}
\usepackage{xcolor}
\usepackage{fancyhdr}
\usepackage{tikz}
\usetikzlibrary{shapes,shapes.geometric,arrows.meta,positioning,decorations.pathreplacing,decorations.pathmorphing,shadows}
\usepackage{pgfplots}
\pgfplotsset{compat=1.18}
\usepgfplotslibrary{fillbetween}
\usepackage{textcomp}

\titleformat{\section}{\large\bfseries}{\thesection.}{0.5em}{}
\titleformat{\subsection}{\normalsize\bfseries}{\thesubsection}{0.5em}{}
\titleformat{\subsubsection}{\normalsize\bfseries\itshape}{\thesubsubsection}{0.5em}{}
\DeclareMathOperator*{\argmax}{arg\,max}
\newcommand{\T}{\mathcal{T}}
\newcommand{\F}{\mathcal{F}}
\newcommand{\w}{\mathbf{w}}

\title{\textbf{Punctuated Equilibria in Artificial Intelligence:}\\[4pt]
\large The Institutional Scaling Law and the Speciation of Sovereign AI}

\author{%
  Mark Baciak$^{1}$, Thomas A. Cellucci$^{1}$, and Deanna M. Falkowski$^{2}$\\[6pt]
  {\small $^{1}$Ekta Inc.\quad $^{2}$Georgetown University}\\
}
\date{}

\begin{document}
\maketitle
\begin{abstract}
\noindent The dominant narrative of artificial intelligence development assumes that progress is continuous and that capability scales monotonically with model size. We challenge both assumptions. Drawing on punctuated equilibrium theory from evolutionary biology, we show that AI development proceeds not through smooth advancement but through extended periods of stasis interrupted by rapid phase transitions that reorganize the competitive landscape. We identify five such eras since 1943 and four epochs within the current Generative AI Era, each initiated by a discontinuous event---from the transformer architecture to the DeepSeek Moment---that rendered the prior paradigm subordinate. To formalize the selection pressures driving these transitions, we develop the Institutional Fitness Manifold, a mathematical framework that evaluates AI systems along four dimensions: capability, institutional trust, affordability, and sovereign compliance. The central result is the Institutional Scaling Law, which proves that institutional fitness is non-monotonic in model scale. Beyond an environment-specific optimum, scaling further \textit{degrades} fitness as trust erosion and cost penalties outweigh marginal capability gains. This directly contradicts classical scaling laws and carries a strong implication: orchestrated systems of smaller, domain-adapted models can mathematically outperform frontier generalists in most institutional deployment environments. We derive formal conditions under which this inversion holds and present supporting empirical evidence spanning frontier laboratory dynamics, post-training alignment evolution, and the rise of sovereign AI as a geopolitical selection pressure.

\medskip
\noindent\textbf{Keywords:} generative AI, large language models, punctuated equilibrium, scaling laws, institutional fitness, capability-trust divergence, sovereign AI, agentic AI, model speciation
\end{abstract}
\vspace{0.5cm}


\section{Introduction}

The field of artificial intelligence has undergone a transformation so rapid and far-reaching over the past decade that conventional historical narratives---typically framed as smooth, monotonic progress---fail to capture the actual dynamics of change. The emergence of generative AI, powered by the transformer architecture \cite{vaswani2017}, has not proceeded as a gradual accumulation of marginal improvements. Rather, the historical record reveals a pattern strikingly reminiscent of \textit{punctuated equilibrium} \cite{gould1972}: extended periods of relative stasis interrupted by sudden, transformative leaps that reorganize the entire landscape of possibilities.

This paper introduces a formal evolutionary taxonomy for the history of AI and, specifically, generative AI. Drawing on frameworks from macroevolution, thermodynamic phase transitions, and complex systems theory, we propose a hierarchical periodization organized as eons, eras, and epochs---matching the geological convention where eras are larger divisions and epochs are subdivisions within them. We formalize our evolutionary framework mathematically by extending the Sustainability Index (SI) of Han et al.\ \cite{han2025} from hardware-level model evaluation to an ecosystem-level Institutional Fitness Manifold, proving that capability and institutional trust can diverge (Capability-Trust Divergence, Theorem~1) and that environmental heterogeneity mathematically necessitates speciation (Proposition~1).

From this framework we obtain the Institutional Scaling Law (Equation~\ref{eq:inst_scaling}), which supersedes classical scaling laws by demonstrating that institutional fitness is non-monotonic in model scale, and that orchestrated systems of domain-specific models can outperform frontier generalists in their native environments (Symbiogenetic Scaling, Equation~\ref{eq:agent}). The formal derivations underlying this framework are presented in Baciak and Cellucci \cite{baciak2026b}; equation numbers in both papers are synchronized to facilitate cross-reference.

Our contribution extends prior work in six critical dimensions: (1) a mathematical framework---the Institutional Fitness Manifold and the Institutional Scaling Law---that formalizes the selection pressures driving AI ecosystem evolution; (2) the Symbiogenetic Scaling correction, proving that domain-specific models tightly coupled to tools, data, and institutional context can exceed the fitness of generalist frontier models; (3) a comprehensive mapping of frontier AI laboratories across geographies, documenting the competitive dynamics driving innovation; (4) a detailed analysis of the accelerating evolution of post-training alignment methods, from RLHF through GRPO and beyond; (5) the introduction of Sovereign AI as a defining phenomenon of the current epoch; (6) an analysis of the DeepSeek Moment of January 2025---a punctuation event that erased \$589 billion in market value, challenged Western AI hegemony, and triggered a new cadence of culturally synchronized model releases culminating in the Lunar New Year 2026 release wave; and (7) an examination of early 2026 developments in agentic orchestration, training democratization, and physical-world agent deployment---developments whose dynamics map onto the framework's mathematical structure while extending the biological analogy from symbiogenesis to niche construction \cite{odlingsmee2003}. We further contextualize model-level advances against empirical findings from the MIT NANDA \textit{State of AI in Business 2025} report \cite{challapally2025}, which documents a stark GenAI Divide: 95\% of enterprise AI pilots produce zero measurable ROI, revealing that institutional absorption of AI capabilities lags far behind the pace of technical innovation.

The remainder of this paper is organized as follows. Section~2 reviews related work. Section~3 presents our formal taxonomy across five eras, its mathematical formalization (Section~3.1), and the Institutional Scaling Law (Section~3.2). Section~4 provides detailed analysis of the Generative AI era including early 2026 developments within the Symbiogenesis epoch (Section~\ref{sec:early2026}), frontier labs, and alignment evolution. Section~5 analyzes the rise of Sovereign AI. Section~6 forecasts future epochs and eras. Section~7 discusses implications, and Section~8 concludes.

\section{Related Work}

The application of evolutionary metaphors to technological change has a rich intellectual history. Loch and Huberman \cite{loch1999} developed a formal punctuated-equilibrium model of technology diffusion. Valverde and Sol\'e \cite{valverde2015} applied phylogenetic network analysis to programming languages, finding bursty innovation patterns. Kaplan et al.\ \cite{kaplan2020} established power-law scaling relationships for neural language models that are strongly analogous to allometric scaling laws in biology. Hoffmann et al.\ \cite{hoffmann2022} refined these with the Chinchilla scaling laws, while Han et al.\ \cite{han2025} demonstrated that these linear scaling assumptions break down for multi-hop reasoning under quantization, revealing a `quantization trap' with significant implications for model compression and deployment trust. Han et al.\ formalized model evaluation through a three-dimensional Sustainability Index (SI) that critically measures Trust, Economic Efficiency, and Environmental Energy---and proved that these dimensions can decouple, with efficiency gains failing to restore trust (Amortization-Trust Decoupling).

We extend their framework in Sections~3.1--3.2 by adding a fourth dimension (sovereign compliance), introducing environment-dependence, proving analogous decoupling and divergence results at the ecosystem level, and deriving a new scaling law that supersedes the classical monotonic formulation. In the multi-agent domain, Lu et al.\ \cite{lu2026} demonstrated that dynamic communication topology routing among specialized LLM agents consistently outperforms fixed communication patterns (+6.2\% avg.\ across benchmarks), providing empirical support for our Symbiogenetic Scaling correction (Section~3.2.1). Yu \cite{yu2026} independently formalized a \textit{Performance Convergence Scaling Law} showing that as frontier models converge toward comparable benchmark performance, orchestration topology---the structural composition of how agents are coordinated---dominates system-level performance over individual model capability, a result structurally parallel to our Convergence-Orchestration Threshold (Equation~\ref{eq:convergence}). Cruzes \cite{cruzes2026} extended the concept of AI sovereignty from data and algorithms to physical infrastructure, demonstrating that practical sovereignty depends on co-design of compute, network, and energy layers---a finding that reinforces our environment-dependent fitness formalization. In the domain-specific deployment space, recent work has empirically demonstrated that small, locally deployed models can deliver sovereign public AI services effectively on modest hardware \cite{sovereign2026services}, providing initial empirical grounding for the Institutional Scaling Law's prediction that $N^*(\varepsilon) \ll N_{\text{frontier}}$ in cost-constrained and sovereignty-weighted environments. Ho et al.\ \cite{ho2025eci} developed the Epoch Capabilities Index (ECI), a composite metric that stitches together dozens of benchmarks via Item Response Theory into a single latent capability scale, enabling cross-model comparison even as individual benchmarks saturate. Crucially for our framework, their acceleration detection methodology---piecewise-linear regression on the frontier ECI series---identified a statistically significant breakpoint near April 2024, with the rate of frontier capability progress nearly doubling from $\sim$8 to $\sim$15 ECI points per year. This empirical finding provides independent, quantitative evidence for the punctuated-equilibrium dynamics we formalize in Section~3.1, and the ECI itself offers a natural operationalization of the Capability index $C(\theta)$ in Definition~1. Kwa et al.\ \cite{metr2025} arrived at a corroborating result through a different methodology: measuring the time horizon of autonomous software engineering tasks, they found a parallel acceleration in 2024 from 7-month to 4-month doubling times. The comprehensive alignment survey by Wang et al.\ \cite{wang2024alignment} catalogues the evolution from RLHF through DPO and beyond, while the survey by Chen et al.\ \cite{chen2025agents} documents self-evolving agent frameworks. The literature on sovereign AI as a strategic imperative argues for institutional control of the entire cognitive stack. None of these works, however, adopt the integrated evolutionary framework---or the formal mathematical apparatus, including the Institutional Scaling Law---we propose here.

\section{A Formal Evolutionary Taxonomy of AI}

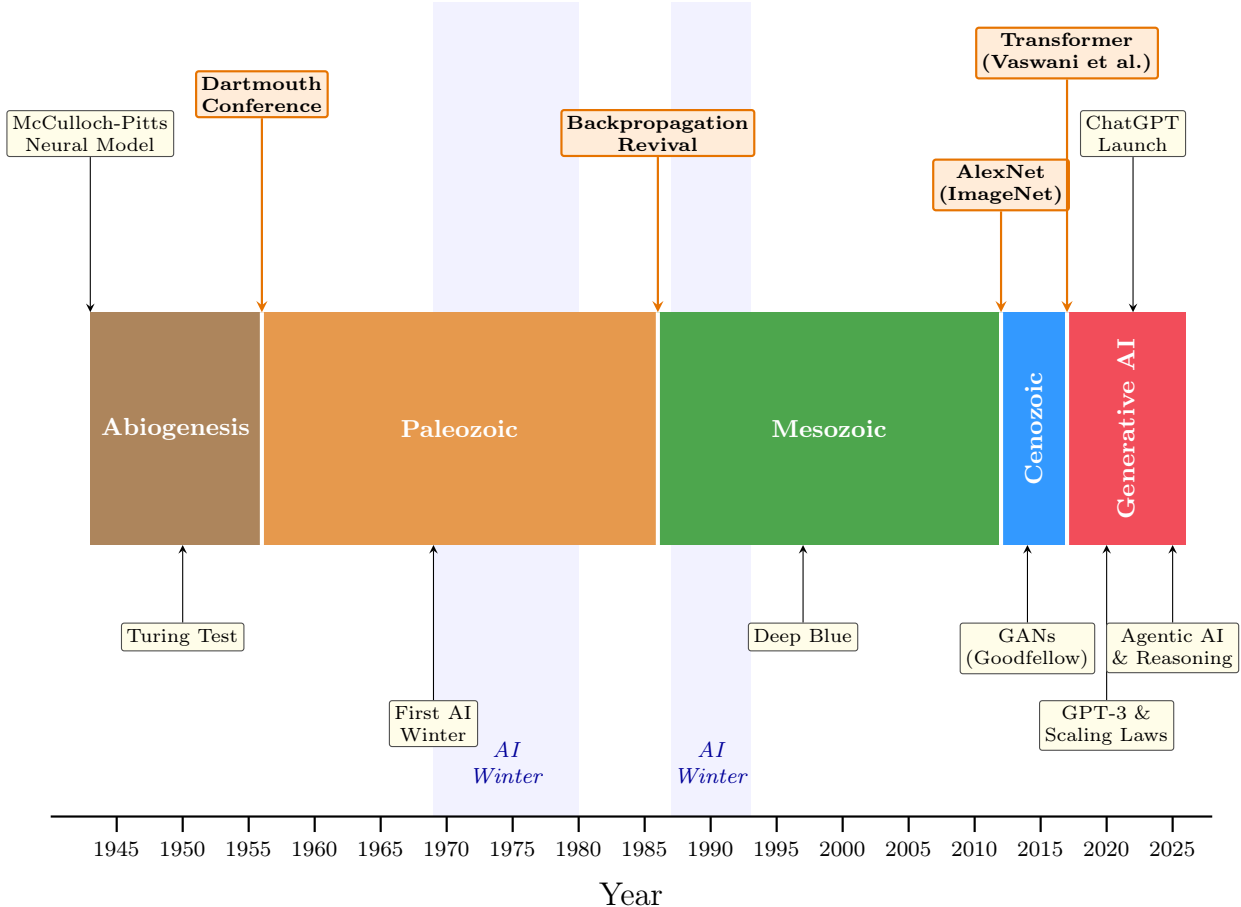
\begin{figure}[H]
\centering
\resizebox{\textwidth}{!}{%
\begin{tikzpicture}[
  x=0.17cm, y=1cm,
  font=\sffamily,
  era box/.style={
    rectangle,
    align=center,
    text=white,
    font=\small\bfseries,
    draw=none,
    inner sep=0pt
  },
  callout/.style={
    rectangle,
    fill=yellow!10,
    draw=black!70,
    align=center,
    font=\scriptsize,
    rounded corners=1pt,
    inner sep=2pt
  },
  transition/.style={
    rectangle,
    fill=orange!15,
    draw=orange!90!black,
    thick,
    align=center,
    font=\scriptsize\bfseries,
    rounded corners=1pt,
    inner sep=2pt
  },
  winter band/.style={
    fill=blue!5,
    draw=none
  }
]

\fill[winter band] (1969, -5.0) rectangle (1980, 5.5);
\node[font=\footnotesize\itshape, text=blue!60!black, align=center] at (1974.5, -4.3) {AI\\Winter};
\fill[winter band] (1987, -5.0) rectangle (1993, 5.5);
\node[font=\footnotesize\itshape, text=blue!60!black, align=center] at (1990, -4.3) {AI\\Winter};

\draw[thick] (1940, -5.0) -- (2028, -5.0);
\foreach \year in {1945, 1950, 1955, 1960, 1965, 1970, 1975, 1980, 1985, 1990, 1995, 2000, 2005, 2010, 2015, 2020, 2025} {
  \draw[thick] (\year, -5.0) -- (\year, -5.2);
  \node[font=\footnotesize, below=0.1cm] at (\year, -5.1) {\year};
}
\node[font=\large] at (1984, -6.0) {Year};

\fill[brown!80!black!80] (1943, -1.5) rectangle (1956, 1.5);
\node[era box, text width=2cm] at (1949.5, 0) {Abiogenesis};

\fill[orange!50!brown!80] (1956, -1.5) rectangle (1986, 1.5);
\node[era box, text width=4cm] at (1971, 0) {Paleozoic};

\fill[green!50!black!70] (1986, -1.5) rectangle (2012, 1.5);
\node[era box, text width=4.5cm] at (1999, 0) {Mesozoic};

\fill[blue!50!cyan!80] (2012, -1.5) rectangle (2017, 1.5);
\node[era box, text width=2.4cm, rotate=90] at (2014.5, 0) {Cenozoic};

\fill[red!70!purple!70] (2017, -1.5) rectangle (2026, 1.5);
\node[era box, text width=2.6cm, rotate=90] at (2021.5, 0) {Generative AI};

\draw[white, line width=1.5pt] (1956, -1.5) -- (1956, 1.5);
\draw[white, line width=1.5pt] (1986, -1.5) -- (1986, 1.5);
\draw[white, line width=1.5pt] (2012, -1.5) -- (2012, 1.5);
\draw[white, line width=1.5pt] (2017, -1.5) -- (2017, 1.5);

\draw[-{stealth}, thin] (1943, 3.5) node[callout, above] {McCulloch-Pitts\\Neural Model} -- (1943, 1.5);
\draw[-{stealth}, thin] (1950, -2.5) node[callout, below] {Turing Test} -- (1950, -1.5);

\draw[-{stealth}, thick, draw=orange!90!black] (1956, 4.0) node[transition, above] {Dartmouth\\Conference} -- (1956, 1.5);

\draw[-{stealth}, thin] (1969, -3.5) node[callout, below] {First AI\\Winter} -- (1969, -1.5);

\draw[-{stealth}, thick, draw=orange!90!black] (1986, 3.5) node[transition, above] {Backpropagation\\Revival} -- (1986, 1.5);
\draw[-{stealth}, thin] (1997, -2.5) node[callout, below] {Deep Blue} -- (1997, -1.5);

\draw[-{stealth}, thick, draw=orange!90!black] (2012, 2.8) node[transition, above] {AlexNet\\(ImageNet)} -- (2012, 1.5);
\draw[-{stealth}, thin] (2014, -2.5) node[callout, below] {GANs\\(Goodfellow)} -- (2014, -1.5);

\draw[-{stealth}, thick, draw=orange!90!black] (2017, 4.5) node[transition, above] {Transformer\\(Vaswani et al.)} -- (2017, 1.5);
\draw[-{stealth}, thin] (2020, -3.5) node[callout, below] {GPT-3 \&\\Scaling Laws} -- (2020, -1.5);
\draw[-{stealth}, thin] (2022, 3.5) node[callout, above] {ChatGPT\\Launch} -- (2022, 1.5);
\draw[-{stealth}, thin] (2025, -2.5) node[callout, below] {Agentic AI\\\& Reasoning} -- (2025, -1.5);

\end{tikzpicture}
}
\caption{Evolutionary Timeline of Artificial Intelligence---From Abiogenesis to the Generative AI Era.}
\label{fig:timeline}
\end{figure}

We propose a hierarchical taxonomy modeled on the geological timescale, following the standard convention: Eon $>$ Era $>$ Epoch, where eras represent the broadest named divisions and epochs are subdivisions within them. In our framework, each of the five eras of AI development is defined by a dominant computational paradigm, and the boundary between eras is marked by a phase transition event---a discontinuous innovation that rendered the preceding paradigm subordinate. Within the current era (Generative AI), we further identify four epochs, each bounded by its own punctuation event: GPT-3 and the demonstration of scaling (June 2020), ChatGPT and mass consumer adoption (November 2022), and the emergence of reasoning-capable agentic systems (September 2024, marked by OpenAI o1). The full taxonomy is summarized in Table~\ref{tab:taxonomy} and Figure~\ref{fig:timeline}.

\begin{table}[H]
\centering
\caption{Evolutionary Taxonomy of AI Development}
\label{tab:taxonomy}
\small
\begin{tabularx}{\textwidth}{@{} >{\raggedright\arraybackslash}p{3.5cm} l >{\raggedright\arraybackslash}X >{\raggedright\arraybackslash}X @{}}
\toprule
\textbf{Era} & \textbf{Period} & \textbf{Defining Characteristics} & \textbf{Phase Transition} \\
\midrule
1. Abiogenesis & 1943--1956 & McCulloch-Pitts, Turing Test, Shannon info theory & Dartmouth Conference (1956) \\
2. Paleozoic (Symbolic) & 1956--1986 & Rule-based AI, expert systems, first AI winter & Expert system collapse \\
3. Mesozoic (Statistical) & 1986--2012 & Backprop revival, SVMs, Bayesian methods, second AI winter, shallow ML & AlexNet / ImageNet (2012) \\
4. Cenozoic & 2012--2017 & CNNs, RNNs/LSTMs, AlphaGo, GANs (2014), VAEs & Transformer (2017) \\
5. Generative AI & 2017--present & Transformers, LLMs, diffusion, RLHF, agentic AI & --- (current era) \\
\bottomrule
\end{tabularx}
\end{table}

While Eras 1--4 establish the foundational trajectory of artificial intelligence, the remainder of this paper focuses its deep analysis entirely on Era~5: the Generative AI Era---where the most consequential evolutionary dynamics are currently unfolding.

\subsection{Mathematical Formalization: The Institutional Fitness Manifold}
\label{sec:manifold}

The evolutionary taxonomy presented above is descriptive. To generate testable predictions and formalize the selection pressures that drive transitions between epochs and eras, we require a mathematical framework. The formal derivations---including all proofs and the derivation of the Institutional Scaling Law---are presented in Baciak and Cellucci \cite{baciak2026b}; here we present the key results and their implications for the evolutionary dynamics documented in this paper. We build on and extend the Sustainability Index (SI) framework of Han et al.\ \cite{han2025}, who formalized model evaluation as a mapping from configuration space to a bounded sustainability vector $\mathbf{v}(\theta) = (T, E, S)^\top$ representing Trust, Economic Efficiency, and Environmental Energy. Their key results---that trust can decouple from efficiency (Amortization-Trust Decoupling, Theorem~4.5) and that apparent optimization can invert into degradation (Scaling Law Divergence, Proposition~3.2)---operate at the level of individual model configurations on specific hardware. We extend this framework from the hardware level to the ecosystem level, introducing environment-dependence to capture the selection pressures---institutional trust, sovereign compliance, cost, and capability---that drive the evolutionary dynamics documented in this paper.

\textbf{Definition 1 (Institutional Fitness Vector).} For any AI system configuration $\theta \in \Theta$ deployed in environment $\varepsilon \in \mathcal{E}$ (where $\mathcal{E}$ indexes the space of deployment contexts: nation-states, regulatory regimes, institutional types), we define the Institutional Fitness Vector:
\begin{equation}
\label{eq:fitness_vec}
f(\theta, \varepsilon) = \bigl(C(\theta),\; \T(\theta,\varepsilon),\; A(\theta),\; \Sigma(\theta,\varepsilon)\bigr)^\top \in [0,1]^4
\end{equation}
where $C(\theta)$ is the Capability index (task performance normalized against the current frontier); $\T(\theta, \varepsilon)$ is the Institutional Trust index, a composite of auditability, behavioral boundedness, and safety verification---critically, this is environment-dependent because different regulatory regimes impose different trust thresholds (where ``trust thresholds'' include but are not limited to regulation or governance of data, re: GDPR for European Systems or cybersecurity considerations re: data leakage for transboundary / cloud based data availability / risk); $A(\theta)$ is the Affordability index (inverse normalized cost-per-query); and $\Sigma(\theta, \varepsilon)$ is the Sovereignty Compliance index (data residency, linguistic attunement, regulatory alignment). This extends Han et al.'s three-dimensional sustainability vector to four dimensions and, crucially, introduces the environment parameter $\varepsilon$ that is absent from their framework. The same model has different fitness in Washington than in Brussels, Beijing, or New Delhi---and this environmental variation is precisely what drives speciation in our evolutionary model.

\textbf{Definition 2 (Scalar Fitness Function).} The scalar institutional fitness of configuration $\theta$ in environment $\varepsilon$ is the inner product:
\begin{equation}
\label{eq:scalar}
F(\theta, \varepsilon) = \w(\varepsilon)^\top \cdot f(\theta, \varepsilon), \quad \sum_i w_i(\varepsilon) = 1
\end{equation}
This directly generalizes Han et al.'s $\text{SI} = \mathbf{w}^\top \mathbf{v}(\theta)$, but the weight vector $\w(\varepsilon)$ now varies by deployment environment. A regulated financial institution in the EU might weight trust and sovereignty heavily ($w_{\T} = 0.35$, $w_\Sigma = 0.30$), while a Silicon Valley startup optimizes primarily for capability and cost ($w_C = 0.45$, $w_A = 0.30$). The heterogeneity of $\w(\varepsilon)$ across environments is the mathematical engine of the adaptive radiation documented in Sections~4--5.

\textbf{Theorem 1 (Capability-Trust Divergence).} The total derivative of institutional fitness with respect to model scale $N$ (parameters) is:
\begin{equation}
\label{eq:divergence}
\frac{\partial F}{\partial N} = w_C \frac{\partial C}{\partial N} + w_{\T} \frac{\partial \T}{\partial N} + w_A \frac{\partial A}{\partial N} + w_\Sigma \frac{\partial \Sigma}{\partial N}
\end{equation}
Under the standard scaling paradigm, one expects $\partial F/\partial N > 0$: bigger models yield higher fitness. However, for regulated environments where $w_{\T}$ is large, the sign flips. Empirically, capability scales approximately as $C(N) \propto N^\alpha$ with $\alpha \approx 0.076$ \cite{kaplan2020}, so $\partial C/\partial N > 0$. But institutional trust degrades with scale: larger models exhibit more latent capabilities \cite{wei2022}, are harder to audit and behaviorally bound, and have more opaque internal dynamics. Therefore $\partial \T/\partial N < 0$ for $N$ beyond some critical threshold $N^*$. When the trust penalty $|w_{\T} \cdot \partial \T/\partial N|$ exceeds the capability gain $|w_C \cdot \partial C/\partial N|$, the gradient $\partial F/\partial N$ flips sign---producing a \textit{Capability-Trust Divergence} directly analogous to Han et al.'s Scaling Law Divergence ($\partial \text{SI}/\partial p > 0$, their Proposition~3.2), but operating at the institutional-ecosystem level rather than the hardware level. The implication is structurally identical: apparent optimization (scaling up) becomes mathematically counterproductive for institutional deployment, just as apparent optimization (quantizing down) is counterproductive for multi-hop reasoning.

\textbf{Theorem 2 (Sequential Trust Degradation).} Institutional trust degrades exponentially with the number of deployment contexts $K$ in which a model operates. Let $\varepsilon_k$ be the probability of a trust-eroding incident (safety failure, data breach, behavioral anomaly, adversarial exploit) in deployment context $k$. The aggregate institutional trust is:
\begin{equation}
\label{eq:trust_deg}
\T_{\text{inst}}(K) = \prod_{k=1}^{K}(1-\varepsilon_k) \approx e^{-\sum \varepsilon_k}, \quad \frac{\partial \T_{\text{inst}}}{\partial K} < 0
\end{equation}
This is structurally identical to Han et al.'s formalization of multi-hop reasoning fragility, where $P(y|x,\theta) = \prod P(h_k|h_{<k},x,\theta)$: errors compound across sequential hops. In our framework, each deployment context is a `hop' across which trust-eroding incidents compound. A model deployed in 700 million weekly interactions (ChatGPT by July 2025) traverses an enormous number of trust-relevant contexts; even a small per-context incident probability $\varepsilon_k$ produces rapid aggregate trust decay. Critically, this degradation satisfies an ecosystem-level analogue of Han et al.'s Amortization-Trust Decoupling: making the model cheaper ($\partial A/\partial \text{cost} > 0$) does not restore trust ($\partial \T/\partial A \approx 0$), just as increasing batch size repairs efficiency but cannot repair reasoning accuracy. Cost collapse and trust erosion are decoupled---which explains why the $30\times$ reduction in API costs documented in Section~6.1 has not prevented the institutional trust deficit described in Section~6.3.

\textbf{Proposition 1 (Speciation via Environmental Isolation).} Let $\theta^*(\varepsilon) = \argmax_\theta F(\theta, \varepsilon)$ be the optimal model configuration for environment $\varepsilon$. If two environments $\varepsilon_1, \varepsilon_2$ have sufficiently different fitness weight vectors, the optimal configurations diverge:
\begin{equation}
\label{eq:speciation}
\| \theta^*(\varepsilon_1) - \theta^*(\varepsilon_2) \| \geq \kappa \cdot \| \w(\varepsilon_1) - \w(\varepsilon_2) \|, \quad \kappa > 0
\end{equation}
where $\kappa > 0$ is a sensitivity constant determined by the convexity of the fitness landscape. This formalizes the central claim of Sections~4--5: sovereign AI is not merely a policy preference but a \textit{mathematical necessity} arising from divergent fitness landscapes. When the EU weights auditability and data sovereignty heavily while China weights capability and state alignment, the optimal models for each environment must diverge---producing the speciation dynamics we observe. The biological analogue is allopatric speciation: geographic isolation (here, regulatory and cultural isolation) drives populations toward distinct fitness optima, eventually producing organisms so specialized that they cannot compete outside their native environment.

\textbf{Definition 3 (Phase Transition Detection).} Define the ecosystem state at time $t$ as the distribution $\Psi(t) = \{(\theta_i, \varepsilon_i, n_i)\}$ where $n_i$ is the deployment frequency of configuration $\theta_i$ in environment $\varepsilon_i$. A phase transition (punctuation event) occurs at time $t^*$ when:
\begin{equation}
\label{eq:phase}
\left.\frac{d}{dt} H(\Psi(t))\right|_{t=t^*} > \lambda_{\text{crit}}, \quad H(\Psi) = -\sum_i p_i \ln p_i
\end{equation}
where $H(\Psi)$ is the Shannon entropy of the configuration-deployment distribution and $\lambda_{\text{crit}}$ is the critical rate threshold. Low $dH/dt$ indicates stasis (the dominant configuration is stable); spikes in $dH/dt$ indicate rapid redistribution of which configurations dominate---exactly the signature of punctuated equilibrium. The ChatGPT launch (November 2022), which restructured the competitive landscape within weeks, and the DeepSeek Moment (January 2025), which invalidated prevailing cost assumptions overnight, both represent episodes where $dH/dt$ dramatically exceeded $\lambda_{\text{crit}}$. This formalization connects the qualitative periodization of Section~3 to a quantitative detection criterion, and provides a prospective tool: monitoring the entropy rate of the deployment distribution offers an early-warning system for forthcoming phase transitions.

\textit{Remark.} The framework above is intentionally conservative. Each definition and theorem admits empirical testing: the weight vectors $\w(\varepsilon)$ can be estimated from procurement data and regulatory filings; the trust degradation rate can be calibrated against documented safety incidents; and the entropy rate of the ecosystem can be computed from model API traffic and deployment surveys. A complementary empirical signal is now available at the capability level: Ho et al.'s \cite{ho2025eci} Epoch Capabilities Index (ECI) provides a unified capability metric whose piecewise-linear frontier admits formal breakpoint detection---their identification of a $\sim$1.9$\times$ acceleration near April 2024 constitutes precisely the kind of rate-change signal that Definition~3 is designed to formalize at the ecosystem level. We leave full empirical calibration to future work, and employ the framework here as a formal structure for organizing and interpreting the evolutionary dynamics documented in the sections that follow.

\subsection{The Institutional Scaling Law}
\label{sec:scaling_law}

The scaling laws of Kaplan et al.\ \cite{kaplan2020} and Hoffmann et al.\ \cite{hoffmann2022} model loss as a power-law function of model size $N$ and data $D$, treating performance in a classical manner as a monotonically improving function of scale. Han et al.\ \cite{han2025} demonstrated that this monotonicity breaks at the hardware level: the `quantization trap' shows that reducing precision paradoxically increases energy consumption and degrades reasoning trust. We now show that an analogous---and more consequential---breakdown occurs at the ecosystem level: institutional fitness is a non-monotonic function of model scale, with an optimal size $N^*(\varepsilon)$ that depends on the deployment environment. We term this the Institutional Scaling Law.

\textbf{Proposition 2 (The Institutional Scaling Law).} The institutional fitness of a model with $N$ parameters, precision $p$, operating in an agentic chain of depth $K$ within deployment environment $\varepsilon$, is:
\begin{align}
\label{eq:inst_scaling}
\F(N&,p,K,\varepsilon) = w_C\!\left[1 - \!\left(\frac{N_c}{N}\right)^{\!\alpha}\right]\nonumber\\
&+ w_{\T}\!\left[\T_0 e^{-\beta N^\gamma}\right] + w_A\!\left[\!\left(\frac{N_r}{N}\right)^{\!\delta}\!\Phi(p)\right] + w_\Sigma \sigma(\varepsilon)
\end{align}
where the four terms correspond to the components of the Institutional Fitness Vector (Definition~1): \textit{Capability} $C(\theta) = 1 - (N_c/N)^\alpha$ follows the classical Kaplan power law, with $\alpha \approx 0.076$; \textit{Trust} $\T(\theta, \varepsilon) = \T_0 \cdot e^{-\beta N^\gamma}$ decays exponentially beyond a critical scale, reflecting the increasing opacity, latent capability proliferation, and auditability failure of large models; \textit{Affordability} $A(\theta) = (N_r/N)^\delta \cdot \Phi(p)$ captures cost-per-query scaling modulated by quantization efficiency; and \textit{Sovereignty} $\Sigma(\theta, \varepsilon) = \sigma(\varepsilon)$ is the environment-specific compliance index. The critical innovation is that the weight vector $\w(\varepsilon) = (w_C, w_{\T}, w_A, w_\Sigma)$ varies by deployment environment, producing fundamentally different fitness landscapes for different institutional contexts.

\begin{figure}[H]
\centering
\begin{tikzpicture}[font=\sffamily]
\begin{axis}[
    width=0.95\textwidth,
    height=9cm,
    xmin=0, xmax=400,
    ymin=0.05, ymax=0.85,
    axis lines=left,
    axis line style={thick},
    xtick={0, 50, 100, 150, 200, 250, 300, 350, 400},
    ytick={0.1, 0.2, 0.3, 0.4, 0.5, 0.6, 0.7, 0.8},
    xlabel={Model Scale $N$ (Billion Parameters)},
    ylabel={Institutional Fitness $\F(N,\varepsilon)$},
    label style={font=\bfseries},
    legend style={at={(0.98,0.98)}, anchor=north east, font=\small,
        draw=black!50, fill=white, fill opacity=0.9, rounded corners=2pt},
    legend cell align={left},
    restrict y to domain=0:1.0,
    grid=major,
    grid style={gray!20},
    every axis plot/.append style={line width=1.2pt}
]

\addplot[
    name path=cap,
    domain=0.5:400,
    samples=200,
    color=gray,
    dashed,
    line width=1pt
] {0.78 * (1 - exp(-0.025*x)) + 0.02};
\addlegendentry{Classical (capability only)}

\addplot[
    name path=startup,
    domain=0.5:400,
    samples=200,
    color=blue!80!black,
    thick
] {min(1, (1 - exp(-0.032*x))) * (0.74 + 0.02 * exp(0.08 * (ln(x/140) - x/140 + 1)))};
\addlegendentry{$\varepsilon_1$: Tech startup ($w_C{=}0.55$)}

\addplot[
    name path=eu,
    domain=0.5:400,
    samples=200,
    color=green!60!black,
    thick
] {min(1, (1 - exp(-0.09*x))) * (0.43 + 0.33 * exp(0.42 * (ln(x/45) - x/45 + 1)))};
\addlegendentry{$\varepsilon_2$: EU regulated institution ($w_{\mathcal{T}}{=}0.40$)}

\addplot[
    name path=sov,
    domain=0.5:400,
    samples=200,
    color=red!70!yellow,
    thick
] {min(1, (1 - exp(-0.15*x))) * (0.24 + 0.16 * exp(0.60 * (ln(x/23) - x/23 + 1)))};
\addlegendentry{$\varepsilon_3$: Sovereign emerging market ($w_A{=}0.58$)}

\addplot[red!10, opacity=0.3] fill between[of=cap and sov, soft clip={domain=29:400}];

\node[circle, fill=red!70!yellow, inner sep=2.5pt] at (axis cs:29, 0.392) {};
\draw[dotted, red!70!yellow, line width=0.8pt] (axis cs:29, 0.392) -- (axis cs:29, 0.05);
\node[above right, font=\small\bfseries, text=red!80!black, fill=white, fill opacity=0.8, text opacity=1, inner sep=1.5pt, rounded corners=1pt] at (axis cs:32, 0.40) {$N^*{\approx}23$B};

\node[circle, fill=green!60!black, inner sep=2.5pt] at (axis cs:54, 0.752) {};
\draw[dotted, green!60!black, line width=0.8pt] (axis cs:54, 0.752) -- (axis cs:54, 0.05);
\node[above, font=\small\bfseries, text=green!60!black, fill=white, fill opacity=0.8, text opacity=1, inner sep=1.5pt, rounded corners=1pt] at (axis cs:54, 0.76) {$N^*{\approx}45$B};

\node[circle, fill=blue!80!black, inner sep=2.5pt] at (axis cs:140, 0.751) {};
\draw[dotted, blue!80!black, line width=0.8pt] (axis cs:140, 0.751) -- (axis cs:140, 0.05);
\node[above, font=\small\bfseries, text=blue!80!black, fill=white, fill opacity=0.8, text opacity=1, inner sep=1.5pt, rounded corners=1pt] at (axis cs:155, 0.76) {$N^*{\approx}140$B};

\node[draw=red!50, fill=red!5, rounded corners, font=\small\itshape, text=red!70!black,
      align=center] at (axis cs:330, 0.15) {Capability-Trust\\Divergence Zone};

\end{axis}
\end{tikzpicture}
\caption{The Institutional Scaling Law. Institutional fitness $\F(N, \varepsilon)$ is non-monotonic: each deployment environment has a distinct optimal scale $N^*(\varepsilon)$. The dashed line shows the classical capability-only view, which increases monotonically. The shaded region marks the Capability-Trust Divergence Zone where scaling up reduces institutional fitness.}
\label{fig:scaling}
\end{figure}
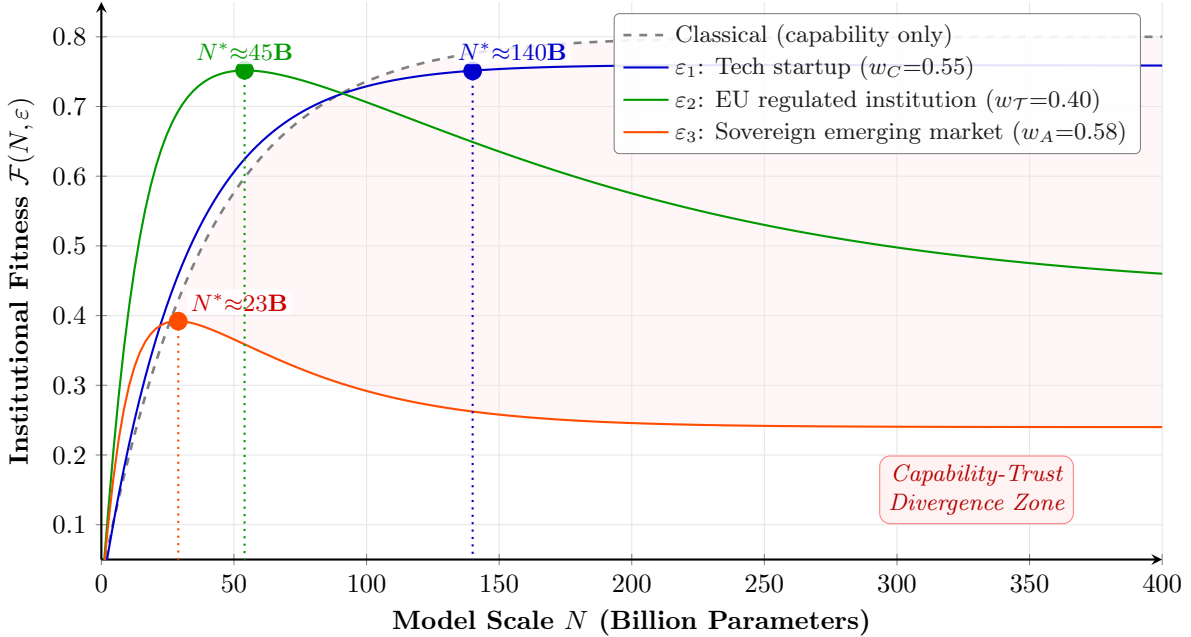

Unlike the classical $L(N, D)$ which decreases monotonically with $N$, institutional fitness $\F(N, p, K, \varepsilon)$ exhibits a non-monotonic profile: it rises as capability increases, reaches a maximum at some environment-specific optimal scale $N^*(\varepsilon)$, and then declines as trust erosion, cost penalties, and sovereignty misalignment outweigh the marginal capability gains. The first-order optimality condition yields the Phase Boundary:
\begin{align}
\label{eq:phase_boundary}
N^*(\varepsilon)&:\; \left.\frac{\partial \F}{\partial N}\right|_{N^*} \!= 0 \;\Rightarrow\nonumber\\
w_C \alpha \frac{N_c^\alpha}{N^{*(\alpha+1)}} &= w_{\T} \beta\gamma \T_0 N^{*(\gamma-1)} e^{-\beta N^{*\gamma}}\nonumber\\
&\quad + w_A \delta \frac{N_r^\delta}{N^{*(\delta+1)}}\Phi(p)
\end{align}
The left side is the marginal capability gain from increasing $N$; the right side is the combined marginal cost in trust erosion and affordability. At $N = N^*(\varepsilon)$, these forces balance. Below $N^*$, capability gains dominate and bigger is better---consistent with the classical scaling paradigm. Above $N^*$, trust and cost penalties dominate and bigger is worse---the Capability-Trust Divergence (Theorem~1). Crucially, $N^*(\varepsilon)$ varies by environment: a Silicon Valley startup optimizing for capability may find $N^* \approx 140$B, while an EU regulated institution weighting trust heavily finds $N^* \approx 45$B, and a sovereign emerging market constrained by cost finds $N^* \approx 23$B. Figure~\ref{fig:scaling} illustrates this divergence.

The quantization-trust interaction term $\Phi(p)$ in Equation~\ref{eq:inst_scaling} directly incorporates Han et al.'s (2025) energy-sustainability framework:
\begin{equation}
\label{eq:quant_trust}
\Phi(p) = \min\!\left(1,\; \frac{\log(1 + \chi_{\text{ref}})}{\log(1 + \chi(p))}\right), \quad \chi(p) = E_q(p) \cdot \gamma_{\text{grid}}
\end{equation}
where $\chi(p)$ is the carbon-adjusted energy score at precision $p$ and $\chi_{\text{ref}}$ is the full-precision baseline. This term couples Han et al.'s hardware-level sustainability analysis to our ecosystem-level fitness: a model that falls into the quantization trap---consuming more energy at lower precision---suffers a direct affordability penalty in the Institutional Scaling Law.

\subsubsection{Symbiogenetic Scaling: The Multi-Agent System Correction}

The Institutional Scaling Law as stated evaluates individual models. But the Symbiogenesis epoch (Section~4.4) is characterized by the fusion of multiple specialized systems into composite agents. This introduces a qualitatively different scaling regime---one where system-level fitness can exceed the fitness of any individual component model. We formalize this through a multi-agent topology correction, motivated by recent work on dynamic agent communication \cite{lu2026}:
\begin{equation}
\label{eq:agent}
\F_{\text{agent}}(N, K, G) = \F(N, p, K, \varepsilon) \cdot \left[1 + \eta \cdot \frac{\rho(G)}{\sqrt{K}}\right]
\end{equation}
where $K$ is the number of specialized agents in the system, $G$ is the communication graph topology, $\rho(G) = |E_{\text{eff}}|/K(K\!-\!1)$ is the effective communication density (the fraction of agent pairs that exchange task-relevant information), and $\eta > 0$ is the orchestration efficiency parameter. The key insight is that domain-specific models tightly coupled to specific tools, trained on system schema and data, and coordinated through adaptive topology routing, can collectively exceed the institutional fitness of a generalist frontier model that has never encountered the deployment environment's tools, data structures, or operational constraints. A 7B model fine-tuned on a hospital's electronic health records, integrated with the hospital's drug interaction database, and coordinated with a 3B radiology specialist and a 2B medical coding agent does not need to compete with GPT-5 on general benchmarks---it needs to outperform GPT-5 in that specific institutional environment, where it has structural advantages in trust (auditable, behaviorally bounded), affordability (runs on commodity hardware), sovereignty (data never leaves the institution), and increasingly, capability on the domain tasks that actually matter.

This produces a Convergence-Orchestration Threshold: the point at which marginal returns from scaling individual models fall below marginal returns from improving system orchestration:
\begin{equation}
\label{eq:convergence}
N_{\text{conv}}:\; \left.\frac{\partial C}{\partial N}\right|_{N_{\text{conv}}} < \mu \;\Longrightarrow\; \frac{\partial \F_{\text{agent}}}{\partial G} > \frac{\partial \F_{\text{agent}}}{\partial N}
\end{equation}
where $\mu$ is a capability saturation threshold. Once individual model capability approaches the frontier ($\partial C/\partial N < \mu$), investment in orchestration topology---how models communicate, divide labor, share context, and coordinate tool use---dominates investment in scale. The ECI data in Figure~\ref{fig:punctuated} offer suggestive evidence: by late 2025, frontier models from multiple labs cluster within a narrow $\sim$10-point ECI band, consistent with the capability convergence regime ($\partial C/\partial N < \mu$) that the threshold predicts. This formalizes a prediction central to our evolutionary framework: the next phase transition in AI will not be triggered by a larger model, but by a better-orchestrated system of specialized models that fuse into a composite intelligence adapted to a specific institutional niche. In biological terms, this is precisely \textit{symbiogenesis} \cite{margulis1967}: the mitochondrion did not outcompete the cell---it merged with it, producing an organism more fit than either ancestor. The domain-specific model system is the mitochondrion of institutional AI.

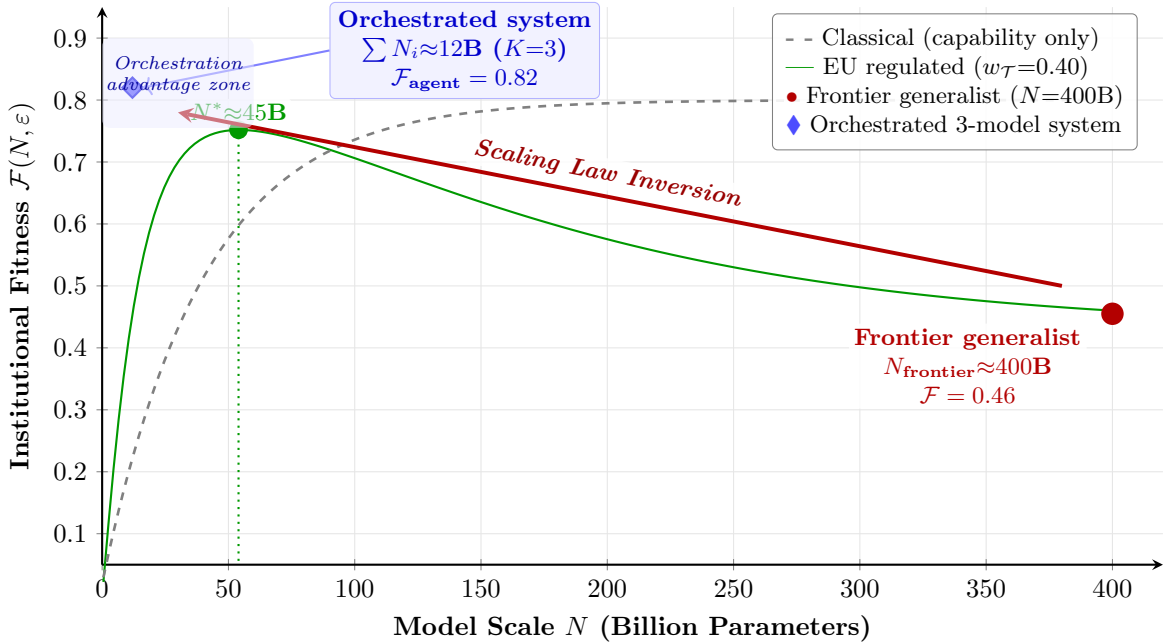
\begin{figure}[H]
\centering
\begin{tikzpicture}[font=\sffamily]
\begin{axis}[
    width=0.95\textwidth,
    height=9cm,
    xmin=0, xmax=420,
    ymin=0.05, ymax=0.95,
    axis lines=left,
    axis line style={thick},
    xtick={0, 50, 100, 150, 200, 250, 300, 350, 400},
    ytick={0.1, 0.2, 0.3, 0.4, 0.5, 0.6, 0.7, 0.8, 0.9},
    xlabel={Model Scale $N$ (Billion Parameters)},
    ylabel={Institutional Fitness $\F(N,\varepsilon)$},
    label style={font=\bfseries},
    restrict y to domain=0:1.0,
    grid=major,
    grid style={gray!20},
    every axis plot/.append style={line width=1.2pt},
    clip=false
]

\addplot[
    domain=0.5:400,
    samples=200,
    color=green!60!black,
    thick,
    name path=eu
] {min(1, (1 - exp(-0.09*x))) * (0.43 + 0.33 * exp(0.42 * (ln(x/45) - x/45 + 1)))};

\addplot[
    domain=0.5:400,
    samples=200,
    color=gray,
    dashed,
    line width=1pt
] {0.78 * (1 - exp(-0.025*x)) + 0.02};

\node[circle, fill=green!60!black, inner sep=2.5pt] at (axis cs:54, 0.752) {};
\draw[dotted, green!60!black, line width=0.8pt] (axis cs:54, 0.752) -- (axis cs:54, 0.05);
\node[above, font=\small\bfseries, text=green!60!black, fill=white, fill opacity=0.85, text opacity=1, inner sep=1.5pt, rounded corners=1pt] at (axis cs:54, 0.76) {$N^*{\approx}45$B};

\node[circle, fill=red!70!black, inner sep=3pt] at (axis cs:400, 0.455) {};
\node[below left, font=\small\bfseries, text=red!70!black, fill=white, fill opacity=0.85, text opacity=1, inner sep=2pt, rounded corners=1pt, align=center] at (axis cs:390, 0.44) {Frontier generalist\\$N_{\text{frontier}}{\approx}400$B\\$\F = 0.46$};

\node[diamond, fill=blue!70, draw=blue!90!black, inner sep=2pt, line width=0.8pt] at (axis cs:12, 0.82) {};
\node[right, font=\small\bfseries, text=blue!80!black, fill=blue!5, draw=blue!30, inner sep=3pt, rounded corners=2pt, align=center] (orchlabel) at (axis cs:90, 0.88) {Orchestrated system\\$\sum N_i{\approx}12$B ($K{=}3$)\\$\F_{\text{agent}} = 0.82$};
\draw[->, blue!50, line width=0.8pt] (orchlabel.west) -- (axis cs:16, 0.82);

\draw[-{stealth}, thick, red!70!black, line width=1.5pt] (axis cs:380, 0.50) -- (axis cs:30, 0.78);
\node[font=\small\bfseries\itshape, text=red!60!black, fill=white, fill opacity=0.85, text opacity=1, inner sep=2pt, rounded corners=1pt, rotate=-11, align=center] at (axis cs:200, 0.68) {Scaling Law Inversion};

\fill[blue!8, opacity=0.5, rounded corners=3pt] (axis cs:0, 0.755) rectangle (axis cs:60, 0.90);
\node[font=\scriptsize\itshape, text=blue!60!black, align=center] at (axis cs:30, 0.84) {Orchestration\\advantage zone};

\node[anchor=north east, draw=black!50, fill=white, fill opacity=0.9, text opacity=1, rounded corners=2pt, inner sep=5pt, font=\small, align=left] at (axis cs:410, 0.94) {%
\textcolor{gray}{\textbf{- -}} Classical (capability only)\\
\textcolor{green!60!black}{\textbf{---}} EU regulated ($w_{\mathcal{T}}{=}0.40$)\\
\textcolor{red!70!black}{$\bullet$} Frontier generalist ($N{=}400$B)\\
\textcolor{blue!70}{$\blacklozenge$} Orchestrated 3-model system};

\end{axis}
\end{tikzpicture}
\caption{Symbiogenetic Scaling Inversion. In a trust-weighted EU deployment environment ($w_{\mathcal{T}} = 0.40$), an orchestrated system of three domain-specific models (7B + 3B + 2B, total 12B parameters) with high communication density $\rho(G)$ achieves institutional fitness $\F_{\text{agent}} = 0.82$---exceeding both the environment's individual-model optimum at $N^* \approx 45$B and the frontier generalist at $N = 400$B ($\F = 0.46$). The star marker denotes the system's aggregate scale; the orchestration multiplier (Equation~\ref{eq:agent}) elevates it above the single-model curve.}
\label{fig:symbioscaling}
\end{figure}

\textbf{Corollary (Scaling Law Inversion for Institutional Deployment).} For institutional deployment environments where $w_{\T}(\varepsilon) + w_\Sigma(\varepsilon) > 0.5$ (trust and sovereignty dominate), there exists a system of $K$ domain-specific models with individual scale $N_i \ll N_{\text{frontier}}$ such that:
$$\F_{\text{agent}}\!\left(\textstyle\sum N_i, K, G, \varepsilon\right) > \F(N_{\text{frontier}}, p, 1, \varepsilon)$$
That is, a system of small specialized models with total parameter count less than the frontier model can achieve higher institutional fitness than the frontier model operating alone---provided the system is adapted to the specific deployment environment. This is the mathematical expression of the evolutionary prediction that it is not the largest organism that survives, but the one best adapted to the selection pressures of its environment. The classical scaling law says bigger is always better. The Institutional Scaling Law says \textit{better-adapted is always better}---and at the ecosystem level, adaptation increasingly means orchestrated specialization rather than undifferentiated scale. Figure~\ref{fig:symbioscaling} illustrates this inversion for a trust-weighted deployment environment.

\section{The Generative AI Era: Deep Analysis}

Having established our five-era taxonomy and its mathematical formalization, we now turn to a detailed analysis of the Generative AI Era---the period from 2017 to the present in which transformer-based systems have come to dominate AI research and increasingly reshape global industry, geopolitics, and culture. We organize this analysis around four distinct epochs within the era, each representing a qualitative shift in the field's dominant dynamics.

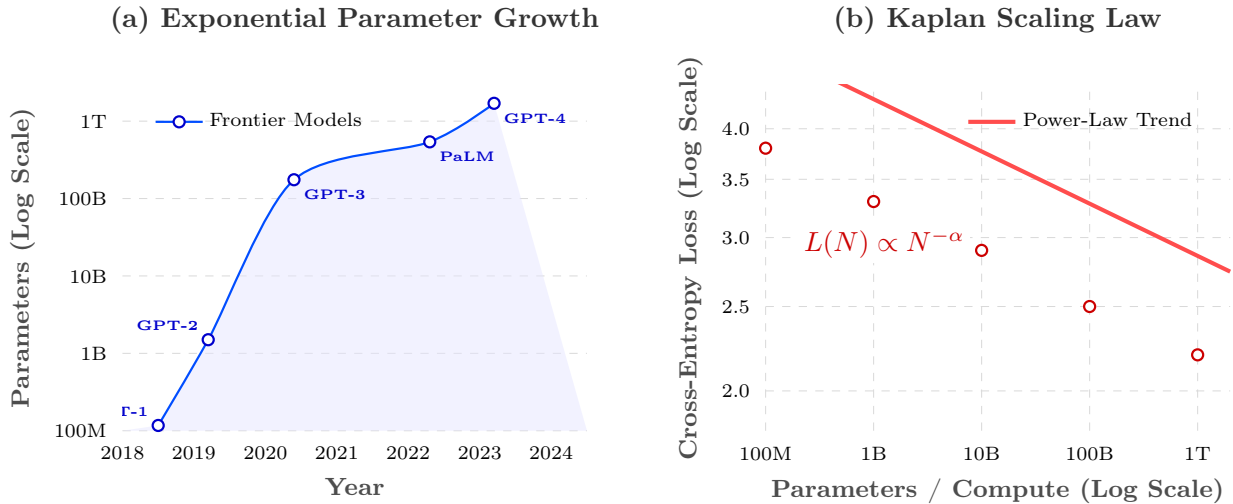
\begin{figure}[H]
\centering
\resizebox{\textwidth}{!}{%
\begin{tikzpicture}[font=\sffamily]

\def\plotwidth{7.5cm}
\def\plotheight{6.0cm}

\begin{axis}[
    name=plotA,
    width=\plotwidth,
    height=\plotheight,
    ymode=log,
    ymin=1e8, ymax=3e12,
    xmin=2018, xmax=2024.5,
    xtick={2018, 2019, 2020, 2021, 2022, 2023, 2024},
    xticklabel style={/pgf/number format/1000 sep=, font=\scriptsize},
    ytick={1e8, 1e9, 1e10, 1e11, 1e12},
    yticklabels={100M, 1B, 10B, 100B, 1T},
    yticklabel style={font=\scriptsize},
    xlabel={Year},
    ylabel={Parameters (Log Scale)},
    label style={font=\small\bfseries, text=black!70},
    title={(a) Exponential Parameter Growth},
    title style={font=\normalsize\bfseries, yshift=2ex, text=black!80},
    axis lines=left,
    axis line style={draw=none},
    tick style={draw=none},
    ymajorgrids=true,
    xmajorgrids=false,
    grid style={line width=.2pt, draw=gray!30, dashed},
    legend style={at={(0.05,0.95)}, anchor=north west, font=\scriptsize, draw=none, fill=none}
]

\addplot[
    name path=A,
    color=blue!70!cyan,
    mark=*,
    mark options={fill=white, draw=blue!80!black, thick, scale=1.0},
    thick,
    smooth
] coordinates {
    (2018.5, 1.17e8)
    (2019.2, 1.5e9)
    (2020.4, 1.75e11)
    (2022.3, 5.4e11)
    (2023.2, 1.7e12)
};
\addlegendentry{Frontier Models}

\path[name path=B] (axis cs:2018,1e8) -- (axis cs:2024.5,1e8);
\addplot[blue!10, opacity=0.5] fill between[of=A and B];

\node[anchor=south east, font=\tiny\bfseries, text=blue!80!black] at (axis cs:2018.5, 1.17e8) {GPT-1};
\node[anchor=south east, font=\tiny\bfseries, text=blue!80!black] at (axis cs:2019.2, 1.5e9) {GPT-2};
\node[anchor=north west, font=\tiny\bfseries, text=blue!80!black] at (axis cs:2020.4, 1.75e11) {GPT-3};
\node[anchor=north west, font=\tiny\bfseries, text=blue!80!black] at (axis cs:2022.3, 5.4e11) {PaLM};
\node[anchor=north west, font=\tiny\bfseries, text=blue!80!black] at (axis cs:2023.2, 1.7e12) {GPT-4};

\end{axis}

\begin{axis}[
    name=plotB,
    at={(plotA.right of south east)},
    xshift=2.0cm,
    width=\plotwidth,
    height=\plotheight,
    xmode=log,
    ymode=log,
    xmin=1e8, xmax=2e12,
    ymin=1.8, ymax=4.5,
    xtick={1e8, 1e9, 1e10, 1e11, 1e12},
    xticklabels={100M, 1B, 10B, 100B, 1T},
    xticklabel style={font=\scriptsize},
    ytick={2, 2.5, 3, 3.5, 4},
    yticklabels={2.0, 2.5, 3.0, 3.5, 4.0},
    yticklabel style={font=\scriptsize},
    xlabel={Parameters / Compute (Log Scale)},
    ylabel={Cross-Entropy Loss (Log Scale)},
    label style={font=\small\bfseries, text=black!70},
    title={(b) Kaplan Scaling Law},
    title style={font=\normalsize\bfseries, yshift=2ex, text=black!80},
    axis lines=left,
    axis line style={draw=none},
    tick style={draw=none},
    ymajorgrids=true,
    xmajorgrids=true,
    grid style={line width=.2pt, draw=gray!30, dashed},
    legend style={at={(0.95,0.95)}, anchor=north east, font=\scriptsize, draw=none, fill=none}
]

\addplot [
    domain=1e8:2e12,
    samples=2,
    color=red!70,
    line width=1.5pt,
] {exp(-0.06 * ln(x) + ln(15))};
\addlegendentry{Power-Law Trend}

\addplot[
    only marks,
    mark=*,
    mark options={fill=white, draw=red!80!black, thick, scale=1.0}
] coordinates {
    (1e8, 3.8)
    (1e9, 3.3)
    (1e10, 2.9)
    (1e11, 2.5)
    (1e12, 2.2)
};

\node[anchor=south west, font=\normalsize\bfseries, text=red!80!black, fill=white, inner sep=2pt] at (axis cs:2e8, 2.8) {$L(N) \propto N^{-\alpha}$};

\end{axis}

\end{tikzpicture}
}
\caption{Allometric Growth in Generative AI---Parameter Scaling and Power-Law Performance. (a) Exponential growth of model parameters over time. (b) Kaplan scaling law: cross-entropy loss as a power law of model size.}
\label{fig:allometric}
\end{figure}

\subsection{Epoch I: Morphogenesis (2017--2020)}

The transformer architecture's emergence in 2017 \cite{vaswani2017} marked the beginning of the Generative AI Era, but its early development was a period of morphogenesis---the establishment of fundamental architectural forms from which all subsequent diversification would flow. This epoch was defined by three scaling milestones that progressively revealed the transformer's potential:

GPT-1 \cite{radford2018} (117M parameters, June 2018) established unsupervised pre-training as a viable paradigm, showing that transformer language models trained on large corpora could be fine-tuned for downstream tasks. BERT \cite{devlin2018} (340M parameters, October 2018) introduced bidirectional pre-training through masked language modeling, achieving state-of-the-art results across 11 NLP benchmarks and catalyzing a `BERTology' research explosion. GPT-2 \cite{radford2019} (1.5B parameters, February 2019) demonstrated coherent long-form text generation and zero-shot task transfer, with OpenAI initially withholding release due to misuse concerns---foreshadowing alignment debates to come.

The epoch's capstone was GPT-3 \cite{brown2020} (175B parameters, June 2020), which demonstrated emergent few-shot learning: the ability to perform tasks specified purely through natural language prompts without any gradient updates. This was a qualitative capability discontinuity---not merely an improvement in degree but a change in kind---and constituted the phase transition into the next epoch. Concurrently, Kaplan et al.\ \cite{kaplan2020} published neural scaling laws establishing that loss scales as a smooth power law with model size, data, and compute (Equation~\ref{eq:kaplan}):
\begin{equation}
\label{eq:kaplan}
L(N) = \left(\frac{N_c}{N}\right)^{\!\alpha_N}\!, \quad \alpha_N \approx 0.076
\end{equation}
converting AI development from empirical art into predictive science. These laws---analogous to the allometric scaling relationships in biology where metabolic rate scales as $B \propto M^{3/4}$---provided the theoretical justification for the massive investment in scale that would define Epoch II (see Figure~\ref{fig:allometric}).

\subsection{Epoch II: Adaptive Radiation (2020--2022)}

GPT-3's demonstration that scaling could produce qualitative capability jumps triggered an explosive diversification across modalities, architectures, and applications---the AI equivalent of the Cambrian Explosion, when major animal body plans appeared within a geologically brief interval. In evolutionary biology, this pattern is called adaptive radiation: a sudden proliferation of species occupying newly available ecological niches. For AI, the niches were computational modalities.

\begin{figure}[H]
\centering
\resizebox{\textwidth}{!}{%
\begin{tikzpicture}[
  x=1cm, y=1cm,
  font=\sffamily,
  >=stealth,
  every node/.style={align=center},
  root/.style={rectangle, rounded corners, draw=black, thick, fill=blue!10, inner sep=8pt, drop shadow, font=\bfseries\large},
  modality/.style={rectangle, rounded corners, draw=black, thick, fill=gray!10, inner sep=6pt, drop shadow, font=\bfseries},
  model/.style={rectangle, rounded corners=3pt, draw=black!70, fill=white, inner sep=4pt, font=\small},
  branch/.style={thick, draw=black!60, -stealth}
]

\foreach \year/\x in {2018/8, 2019/10, 2020/12, 2021/14, 2022/16, 2023/18, 2024/20.5, 2025/23} {
  \draw[thick, dashed, gray!50] (\x, 6.5) -- (\x, -6.0);
  \node[above, font=\bfseries\color{gray!80!black}] at (\x, 6.5) {\year};
}

\node[root] (transformer) at (0, 0) {Transformer\\Architecture\\(2017)};

\node[modality, fill=red!15] (text) at (4.5, 5.0) {Text (LLMs)};
\node[modality, fill=orange!15] (code) at (4.5, 3.6) {Code};
\node[modality, fill=yellow!15] (image) at (4.5, 2.2) {Image};
\node[modality, fill=green!15] (audio) at (4.5, 0.8) {Audio};
\node[modality, fill=cyan!15] (video) at (4.5, -0.6) {Video};
\node[modality, fill=blue!15] (science) at (4.5, -2.0) {Science / Bio};
\node[modality, fill=purple!15] (multimodal) at (4.5, -3.4) {Multi-modal};
\node[modality, fill=magenta!15] (reasoning) at (4.5, -4.8) {Reasoning / Agents};

\foreach \mod in {text, code, image, audio, video, science, multimodal, reasoning} {
  \draw[branch, shorten <=2pt, shorten >=2pt] (transformer.east) to[out=0, in=180] (\mod.west);
}

\node[model] (gpt1) at (8, 5.0) {GPT-1};
\node[model] (gpt3) at (12, 5.0) {GPT-3};
\node[model] (chatgpt) at (16, 5.0) {ChatGPT};
\node[model] (llama) at (18, 5.0) {LLaMA};
\draw[branch] (text.east) -- (gpt1.west);
\draw[branch] (gpt1.east) -- (gpt3.west);
\draw[branch] (gpt3.east) -- (chatgpt.west);
\draw[branch] (chatgpt.east) -- (llama.west);

\node[model] (codex) at (14, 3.6) {Codex};
\node[model] (starcoder) at (18, 3.6) {StarCoder};
\draw[branch] (code.east) -- (codex.west);
\draw[branch] (codex.east) -- (starcoder.west);

\node[model] (dalle) at (14, 2.2) {DALL-E};
\node[model] (sd) at (16, 2.2) {Stable Diff.};
\node[model] (midjourney) at (20.5, 2.2) {Midjourney v6};
\draw[branch] (image.east) -- (dalle.west);
\draw[branch] (dalle.east) -- (sd.west);
\draw[branch] (sd.east) -- (midjourney.west);

\node[model] (whisper) at (16, 0.8) {Whisper};
\node[model] (musiclm) at (18, 0.8) {MusicLM};
\node[model] (suno) at (20.5, 0.8) {Suno v3};
\draw[branch] (audio.east) -- (whisper.west);
\draw[branch] (whisper.east) -- (musiclm.west);
\draw[branch] (musiclm.east) -- (suno.west);

\node[model] (gen2) at (18, -0.6) {Gen-2};
\node[model] (sora) at (20.5, -0.6) {Sora};
\node[model] (veo) at (22.5, -0.6) {Veo};
\draw[branch] (video.east) -- (gen2.west);
\draw[branch] (gen2.east) -- (sora.west);
\draw[branch] (sora.east) -- (veo.west);

\node[model] (alphafold) at (12, -2.0) {AlphaFold 2};
\node[model] (esm) at (16, -2.0) {ESMFold};
\node[model] (alphafold3) at (20.5, -2.0) {AlphaFold 3};
\draw[branch] (science.east) -- (alphafold.west);
\draw[branch] (alphafold.east) -- (esm.west);
\draw[branch] (esm.east) -- (alphafold3.west);

\node[model] (gpt4) at (18, -3.4) {GPT-4};
\node[model] (gemini) at (20.5, -3.4) {Gemini 1.5};
\node[model] (claude) at (22.5, -3.4) {Claude 3.5};
\draw[branch] (multimodal.east) -- (gpt4.west);
\draw[branch] (gpt4.east) -- (gemini.west);
\draw[branch] (gemini.east) -- (claude.west);

\node[model] (autogpt) at (18, -4.8) {AutoGPT};
\node[model] (o1) at (20.5, -4.8) {OpenAI o1};
\node[model] (deepseek) at (23, -4.8) {DeepSeek-R1};
\draw[branch] (reasoning.east) -- (autogpt.west);
\draw[branch] (autogpt.east) -- (o1.west);
\draw[branch] (o1.east) -- (deepseek.west);

\end{tikzpicture}
}
\caption{Taxonomic Diversification of Generative AI Models Across Modalities (2018--2025). The cladogram shows rapid branching from the common transformer ancestor into text, code, image, video, audio, science, multi-modal, and reasoning/agent lineages across a temporal axis.}
\label{fig:diversification}
\end{figure}
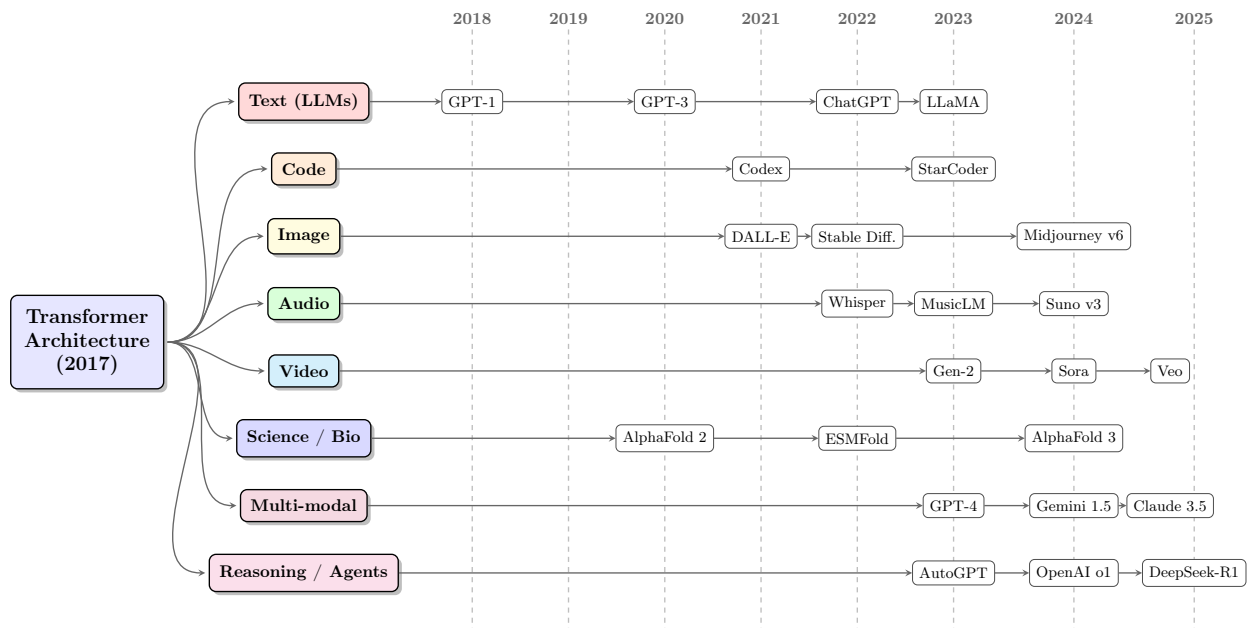

DALL$\cdot$E \cite{ramesh2021} (January 2021) applied the transformer to text-to-image generation, colonizing the visual modality. Codex \cite{chen2021code} (August 2021) specialized in code generation, producing GitHub Copilot. PaLM \cite{chowdhery2022} (April 2022) scaled to 540B parameters, achieving breakthrough performance on reasoning tasks. Stable Diffusion \cite{rombach2022} (August 2022) democratized image generation through open-source release, enabling a Cambrian explosion of creative tools. AlphaFold \cite{jumper2021} (July 2021) solved the 50-year protein folding problem, demonstrating that AI could produce genuine scientific breakthroughs. The period also saw Goodfellow's GAN framework \cite{goodfellow2014} reach maturity in image synthesis, while diffusion models emerged as a powerful alternative generative paradigm (see Figure~\ref{fig:diversification}).

Hoffmann et al.\ \cite{hoffmann2022} refined scaling laws with the Chinchilla paper (March 2022), demonstrating that many large models were significantly undertrained relative to their parameter count---implying that the field had been over-investing in model size and under-investing in training data. This insight triggered a recalibration: subsequent models (Llama, Mistral) achieved comparable performance at smaller scales through better data curation and training efficiency, a finding that would prove prescient for the shift toward smaller, domain-specific models described in Section~6.

By 2022, transformer-based systems had colonized every major computational modality---text, code, images, audio, video, protein structures, and mathematical reasoning---establishing the architectural monoculture that would define the subsequent epoch. The transformer had become the universal substrate of generative AI, the singular class of organism capable of adapting to any informational niche.

\subsection{Epoch III: The Great Expansion (November 2022--2024)}

If Epoch II established that transformers could diversify across modalities, Epoch III demonstrated that they could scale across society. ChatGPT's release on November 30, 2022 constituted a speciation event of unprecedented speed---reaching 100 million users within two months, the fastest consumer technology adoption in history. This was not primarily a technical advance (ChatGPT used GPT-3.5, which was architecturally iterative): it was a deployment innovation---the packaging of a language model as a conversational interface that any human could use without technical expertise. The key technical innovation was RLHF \cite{ouyang2022}, which demonstrated that post-training alignment via human feedback could produce models that were significantly more useful, honest, and safe than their base pre-trained versions---and that a 1.3B InstructGPT model could be preferred by human evaluators over the 175B GPT-3 on instruction-following tasks. This single result challenged the monotonic scaling assumption: alignment quality could substitute for raw scale.

\begin{figure}[H]
\centering
\resizebox{\textwidth}{!}{%
\begin{tikzpicture}[font=\sffamily]
\begin{axis}[
    width=18cm,
    height=12cm,
    xmin=2022.9, xmax=2026.35,
    ymin=72, ymax=165,
    axis lines=left,
    axis line style={thick},
    xtick={2023.0, 2023.5, 2024.0, 2024.5, 2025.0, 2025.5, 2026.0},
    xticklabels={Jan '23, Jul '23, Jan '24, Jul '24, Jan '25, Jul '25, Jan '26},
    xticklabel style={font=\normalsize},
    ytick={80, 90, 100, 110, 120, 130, 140, 150, 160},
    yticklabel style={font=\normalsize},
    ylabel={Epoch Capabilities Index (ECI)},
    label style={font=\large\bfseries},
    grid=major,
    grid style={gray!15},
    clip=false
]

\fill[blue!4, opacity=0.5] (axis cs:2022.9,72) rectangle (axis cs:2024.27,163);
\fill[red!4, opacity=0.5] (axis cs:2024.27,72) rectangle (axis cs:2026.35,163);

\addplot[only marks, mark=*, mark size=2pt, color=gray!45] coordinates {
    (2023.15, 95.1) (2023.15, 99.7) (2023.15, 107.4) (2023.15, 110.4)
    (2023.22, 77.2) (2023.34, 92.7) (2023.45, 122.1)
    (2023.53, 97.8) (2023.53, 105.9) (2023.53, 114.6) (2023.54, 119.7)
    (2023.69, 112.5) (2023.70, 89.3) (2023.74, 113.0)
    (2023.83, 96.2) (2023.89, 118.4) (2023.89, 117.9)
    (2023.95, 118.7) (2023.95, 107.6)
    (2024.13, 117.2) (2024.14, 93.5) (2024.14, 111.8) (2024.15, 121.2)
    (2024.16, 120.1) (2024.16, 117.8) (2024.29, 121.4)
    (2024.29, 116.9) (2024.29, 119.6) (2024.31, 123.0)
    (2024.43, 125.7) (2024.48, 119.8) (2024.48, 122.6)
    (2024.55, 127.1) (2024.56, 115.9) (2024.56, 125.5) (2024.56, 128.0)
    (2024.70, 135.7) (2024.72, 129.5) (2024.81, 134.3) (2024.81, 127.5)
    (2024.94, 127.5) (2024.95, 131.2) (2024.98, 133.2)
    (2025.05, 139.9) (2025.07, 133.5) (2025.08, 141.6)
    (2025.10, 135.7) (2025.15, 142.2) (2025.16, 137.6) (2025.20, 131.1)
    (2025.23, 137.2) (2025.26, 130.4) (2025.26, 133.2) (2025.27, 139.2)
    (2025.29, 135.8) (2025.29, 137.9) (2025.29, 146.3)
    (2025.33, 139.7) (2025.35, 143.5) (2025.39, 142.9) (2025.39, 143.7)
    (2025.46, 146.5) (2025.52, 147.8) (2025.60, 145.0)
    (2025.66, 144.3) (2025.72, 147.0) (2025.75, 146.8) (2025.77, 150.3)
    (2025.85, 149.7) (2025.91, 149.7) (2025.93, 145.7)
    (2026.07, 148.0) (2026.10, 153.0) (2026.13, 156.0)
};

\addplot[thick, color=red!70!black, mark=*, mark size=3pt, line width=1.5pt] coordinates {
    (2023.20, 126.2) (2024.16, 127.0) (2024.27, 127.7) (2024.47, 130.0)
    (2024.70, 136.5) (2024.96, 143.0) (2025.25, 145.1) (2025.29, 147.2)
    (2025.44, 148.2) (2025.60, 150.0) (2025.77, 150.3) (2025.88, 153.4)
    (2025.95, 153.8) (2026.10, 155.1) (2026.14, 157.1) (2026.18, 158.2)
};

\addplot[dashed, color=blue!50!gray, line width=1.2pt, domain=2023.0:2024.27, samples=2]
    {120 + 8.3*(x - 2024.27)};

\addplot[dashed, color=red!50!gray, line width=1.2pt, domain=2024.27:2026.35, samples=2]
    {120 + 15.5*(x - 2024.27)};

\draw[dashed, orange!80!black, line width=1.2pt] (axis cs:2024.27, 72) -- (axis cs:2024.27, 163);
\node[font=\small\bfseries\itshape, text=orange!70!black, align=center, fill=white, fill opacity=0.85, text opacity=1, inner sep=2pt, rounded corners=1pt] at (axis cs:2024.27, 162) {Capability acceleration\\breakpoint (Apr 2024)};

\node[font=\small\bfseries, text=blue!50!gray, fill=white, fill opacity=0.85, text opacity=1, inner sep=2pt, rounded corners=1pt] at (axis cs:2023.55, 108) {$\sim$8 ECI/yr};
\node[font=\small\bfseries, text=red!50!gray, fill=white, fill opacity=0.85, text opacity=1, inner sep=2pt, rounded corners=1pt] at (axis cs:2025.55, 138) {$\sim$16 ECI/yr};

\node[font=\scriptsize\bfseries, text=red!70!black, anchor=south west, fill=white, fill opacity=0.85, text opacity=1, inner sep=1.5pt, rounded corners=1pt] at (axis cs:2023.23, 127) {GPT-4};
\node[font=\scriptsize\bfseries, text=red!70!black, anchor=south west, fill=white, fill opacity=0.85, text opacity=1, inner sep=1.5pt, rounded corners=1pt] at (axis cs:2024.73, 143.5) {o1};
\node[font=\scriptsize\bfseries, text=blue!70!black, anchor=south east, fill=white, fill opacity=0.85, text opacity=1, inner sep=1.5pt, rounded corners=1pt] at (axis cs:2025.03, 140.5) {DeepSeek-R1};
\node[font=\scriptsize\bfseries, text=red!70!black, anchor=south west, fill=white, fill opacity=0.85, text opacity=1, inner sep=1.5pt, rounded corners=1pt] at (axis cs:2025.62, 150.5) {GPT-5};
\node[font=\scriptsize\bfseries, text=red!70!black, anchor=south west, fill=white, fill opacity=0.85, text opacity=1, inner sep=1.5pt, rounded corners=1pt] at (axis cs:2026.00, 158.5) {GPT-5.4 Pro};

\node[anchor=north west, draw=black!50, fill=white, fill opacity=0.9, text opacity=1, rounded corners=2pt, inner sep=5pt, font=\small, align=left] at (axis cs:2022.95, 151) {%
\textcolor{gray!50}{$\bullet$} Individual models (ECI score)\\
\textcolor{red!70!black}{\textbf{---}} Frontier progression (record-setters)\\
\textcolor{blue!50!gray}{\textbf{- -}} Pre-breakpoint trend ($\sim$8 ECI/yr)\\
\textcolor{red!50!gray}{\textbf{- -}} Post-breakpoint trend ($\sim$16 ECI/yr)};

\end{axis}
\end{tikzpicture}
}
\caption{AI Capability Acceleration---Epoch Capabilities Index (2023--2026). Each point represents a model's composite capability score on the Epoch Capabilities Index (ECI), which unifies benchmarks into a single scale using Item Response Theory \cite{ho2025eci}. The red line traces the frontier (record-setting models). Dashed lines show piecewise linear trends: capability progress nearly doubled in rate at the April 2024 breakpoint ($\sim$8 ECI/year $\to$ $\sim$16 ECI/year), several months before the release of reasoning-augmented models (o1, September 2024). Kwa et al.\ \cite{metr2025} independently corroborate the acceleration timing via METR task-horizon data. This empirical acceleration pattern provides quantitative evidence for the punctuated equilibrium dynamic described in this paper.}
\label{fig:punctuated}
\end{figure}
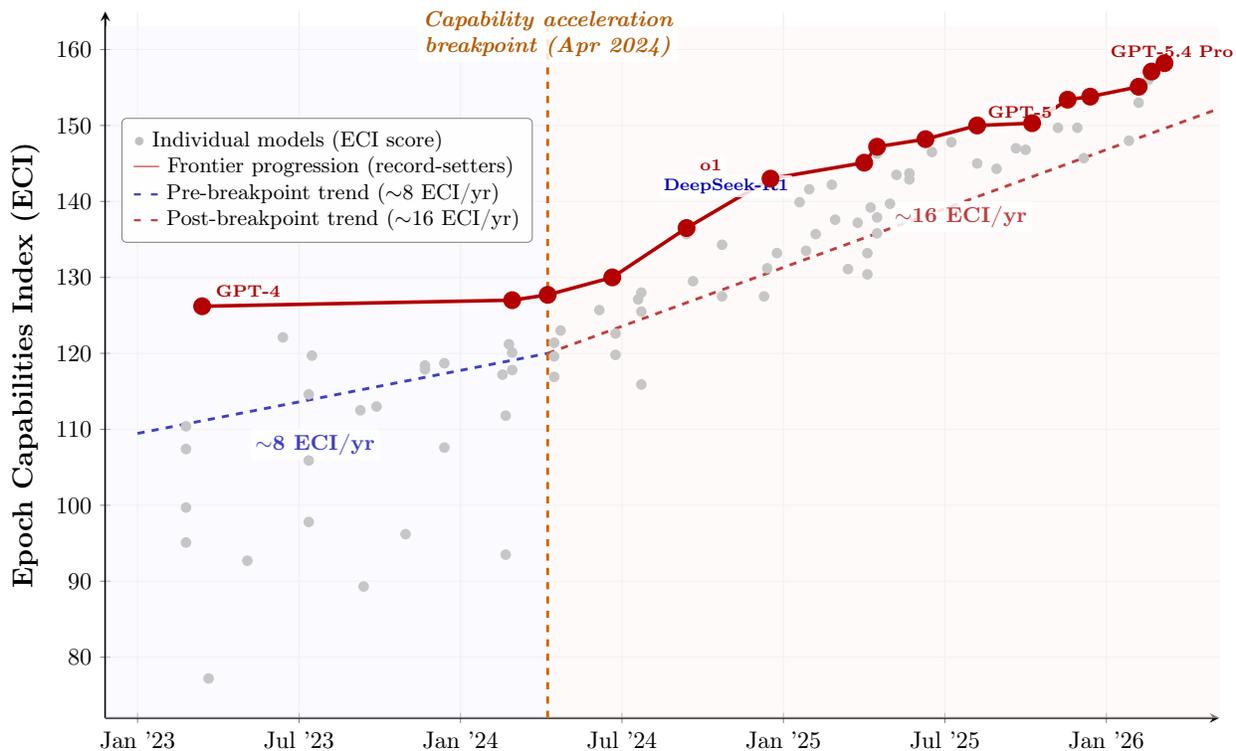

GPT-4 \cite{openai2023} (March 2023) demonstrated multimodal reasoning. Google launched Gemini (December 2023). Anthropic released Claude 2 and Claude 3. Enterprise adoption surged: McKinsey's 2025 survey found AI adoption rising from approximately 6\% (2023) to 30\% (2025). The accelerating frequency of these phase transitions---from decades-long gaps between early eras to year-scale intervals within the Generative AI era---is the defining signature of punctuated equilibrium in AI development. This acceleration is quantified in Figure~\ref{fig:punctuated}, which plots model-level scores on the Epoch Capabilities Index (ECI)---a composite scale that unifies benchmarks via Item Response Theory \cite{ho2025eci}---against time. The frontier progression reveals a striking structural break: capability advanced at roughly 8 ECI points per year through early 2024, then nearly doubled to approximately 16 ECI points per year after an April 2024 breakpoint. Ho et al.\ \cite{ho2025eci} note that this breakpoint precedes the release of reasoning-augmented models such as OpenAI's o1 (September 2024) by several months, suggesting that the acceleration reflects a broader shift---increased investment in reinforcement learning, richer post-training regimes, and the maturation of scaling infrastructure---of which reasoning models are the most visible but not sole manifestation. Kwa et al.\ \cite{metr2025} independently corroborate the timing through a different lens: the doubling time of autonomous software engineering task horizons accelerated from 7 months to 4 months in 2024, yielding a $\sim$1.75$\times$ acceleration consistent with the ECI's $\sim$1.9$\times$ finding. The piecewise acceleration is precisely the empirical pattern that a punctuated-equilibrium model predicts---long intervals of incremental improvement interrupted by rapid, phase-transition-like leaps---and it bridges the qualitative narrative of this section with the quantitative framework developed in Sections~3.1--3.2.

\subsection{Epoch IV: Symbiogenesis (September 2024--Present)}

OpenAI's o1 (September 2024) demonstrated extended chain-of-thought reasoning with deliberative alignment: the model could `think' through complex problems in a multi-step internal process before generating a response. This signaled the transition from static response generation to dynamic, multi-step cognitive architectures capable of planning, reflection, and self-correction. The epoch's defining biological analogy is \textit{symbiogenesis} \cite{margulis1967}: language models, code interpreters, search engines, memory systems, and tool-use frameworks fusing into integrated agentic AI systems---composite organisms more capable than any individual component.

Key developments defining this epoch include: chain-of-thought and extended reasoning (OpenAI o1/o3, DeepSeek R1, Anthropic Claude 3.5 Sonnet); persistent memory and extended context (Gemini 1.5 Pro with 1M+ token context window, GPT-4.1 with 1M tokens); multi-agent orchestration frameworks enabling specialized models to coordinate on complex tasks; and the emergence of AI systems that autonomously write code, browse the web, execute multi-step research workflows, and coordinate with other AI systems (Claude Code, GPT-5 Codex, Devin, Manus). By July 2025, ChatGPT alone reached approximately 700 million weekly active users \cite{cnbc2025}, making it the fastest-growing consumer technology platform in history.

The January 2025 release of DeepSeek R1---achieving performance comparable to OpenAI's o1 at a fraction of the cost and released under the MIT License---constituted the defining punctuation event of the epoch, sparking what became known as the `DeepSeek Moment' (analyzed in detail in Section~4.5.1).

\subsubsection{Orchestration, Democratization, and Niche Construction (Early 2026)}
\label{sec:early2026}

Three developments in early 2026 illustrate the dynamics that the Institutional Scaling Law (Equation~\ref{eq:inst_scaling}) and the Symbiogenetic Scaling correction (Equation~\ref{eq:agent}) formalize. We present them not as confirmation of a prediction but as observable instantiations of the mathematical structure developed in Sections~3.1--3.2, and note where the evidence is partial or where the framework's applicability requires qualification.

\textbf{Agentic orchestration as a competitive strategy.} Anthropic's Claude Code \cite{anthropic2026agent} exemplifies the system-level composition that Equation~\ref{eq:agent} describes. Claude Code operates as an agentic harness in which a frontier language model is orchestrated with file-system access, shell execution, sub-agent delegation, web search, and persistent project context into a composite system that chains an average of 21.2 independent tool calls per task without human intervention \cite{anthropic2026autonomy}. Anthropic's own telemetry shows that the 99.9th percentile autonomous turn duration nearly doubled between October 2025 and January 2026---from under 25 minutes to over 45 minutes---and that this increase was smooth across model releases, suggesting the gains derive from orchestration maturity and user trust rather than from increased model scale alone.

In the language of Equation~\ref{eq:agent}, the competitive advantage of Claude Code maps to the $\rho(G)/\sqrt{K}$ term: the effective communication density among specialized sub-processes (code execution, file reading, test verification, planning) produces system-level fitness that exceeds any single inference call to the same underlying model. The Convergence-Orchestration Threshold (Equation~\ref{eq:convergence}) provides the formal structure: once base model capability approaches the frontier ($\partial C/\partial N < \mu$), marginal returns from improving orchestration topology dominate marginal returns from scaling the model further. Yu \cite{yu2026} arrived at a structurally parallel result independently, formalizing a Performance Convergence Scaling Law showing that as frontier models cluster within a narrow benchmark range, orchestration topology becomes the primary lever for system-level performance gains. Claude Code's architecture reflects exactly this calculus---Anthropic invested in orchestration quality, sub-agent coordination, and tool integration rather than simply releasing a larger model.

A necessary qualification: Claude Code's base model is itself frontier-scale, not the small domain-specific model that the Scaling Law Inversion Corollary describes. What the system instantiates is the \textit{orchestration} component of Symbiogenetic Scaling---the demonstration that system-level composition outperforms raw model inference---rather than the full domain-specific specialization thesis. The stronger claim---that orchestrated systems of \textit{small} models outperform frontier generalists in institutional niches---remains a structural implication of the framework whose full empirical realization is still emerging.

\textbf{Democratization of training infrastructure.} Karpathy's \textit{microgpt} project (February 2026) \cite{karpathy2026microgpt} distills the complete algorithmic content of GPT training and inference---tokenizer, autograd engine, transformer architecture, optimizer, training loop, and inference loop---into 243 lines of dependency-free Python. His companion project \textit{nanochat} \cite{karpathy2026nanochat} demonstrates end-to-end reproduction of GPT-2 (124M) on a single 8$\times$H100 node in approximately two hours, with AI agents contributing 110 code optimizations in 12 hours without human intervention.

These projects bear on the Institutional Scaling Law through its environmental heterogeneity. If the optimal model scale $N^*(\varepsilon)$ for many institutional environments is substantially below the frontier---our framework estimates $N^* \approx 23$B--$45$B for cost-constrained and trust-weighted environments---then the practical relevance of this result depends on whether institutions can actually train and deploy at those scales. Karpathy's work addresses the \textit{supply side} of this dynamic: it reduces the infrastructure, expertise, and cost barriers to training sub-frontier models, making the local optima identified by the Institutional Scaling Law accessible to a broader population of institutional actors. The interaction with sovereign AI dynamics (Section~5) is direct: national programs pursuing ``frugal, sovereign, and scalable'' AI strategies (Section~5.4) require exactly this kind of accessible training infrastructure to reach their environment-specific $N^*(\varepsilon)$. Recent empirical work has demonstrated that small, locally deployed language models can deliver sovereign public AI services---citizen-facing conversational systems for government agencies---effectively on modest computational and financial resources while maintaining cultural and digital autonomy \cite{sovereign2026services}, providing initial empirical grounding for the claim that institutional environments weighted toward sovereignty and affordability can be well-served at scales far below the frontier.

\textbf{Agentic systems in physical environments.} The OpenClaw framework \cite{openclaw2026} (247,000+ GitHub stars by March 2026)---an open-source autonomous agent that orchestrates language models with messaging platforms, skill systems, file management, and increasingly, physical robotics---illustrates a dynamic that extends symbiogenesis beyond the digital domain. OpenClaw agents have been paired with robotic hardware (e.g., Unitree G1 humanoid robots), creating composite systems where a language model reasons about tasks while physical actuators execute them in the material world \cite{openclaw2026robotics}. The peaq Robotics SDK integration enables robots to receive and execute reusable agent skills, establishing bidirectional flows between digital reasoning and physical manipulation.

In the framework's terms, the physical deployment environment represents a radically different fitness weight vector $\w(\varepsilon)$ from text-based benchmarks: a generalist LLM has near-zero fitness in a manipulation task where the selection pressures are latency, physical safety, and sensorimotor coupling. The Speciation Proposition (Equation~\ref{eq:speciation}) applies directly: the optimal configuration for a physical deployment environment must diverge from the text-optimized frontier. What OpenClaw demonstrates is that the \textit{symbiogenetic architecture}---specialized components fused into a composite system---can bridge this gap, creating organisms (in our evolutionary vocabulary) that inhabit niches no individual component could occupy alone.

\textbf{From symbiogenesis to niche construction.} Collectively, these developments exhibit a dynamic that extends beyond the fusion metaphor of classical symbiogenesis. In evolutionary biology, \textit{niche construction} \cite{odlingsmee2003} describes organisms that modify the selective environment in which they and their descendants will be evaluated---beavers building dams, earthworms altering soil chemistry. The agentic systems emerging in early 2026 are niche constructors: Claude Code reshapes the codebases it operates on, creating new affordances for its own future operation; OpenClaw agents deploy skills to robots that then expose new capabilities back to agents; Karpathy's AI-assisted training optimizations modify the very infrastructure that will produce the next generation of models. The system is not merely \textit{adapted to} its environment---it is actively \textit{modifying} that environment. Figure~\ref{fig:nicheconstruction} illustrates this transition.

\definecolor{transitioncol}{RGB}{210, 100, 50}
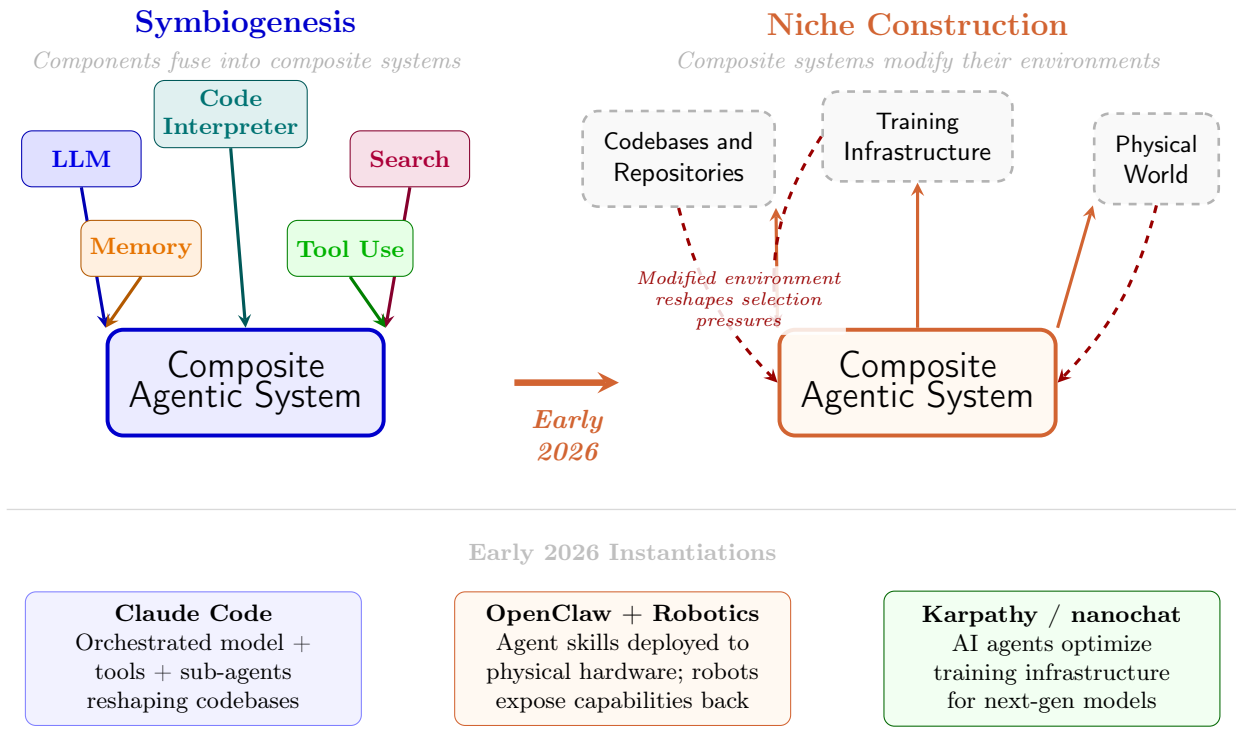
\begin{figure}[H]
\centering
\resizebox{\textwidth}{!}{%
\begin{tikzpicture}[font=\sffamily,
  phaselbl/.style={font=\large\bfseries},
  envnode/.style={rectangle, rounded corners=6pt, draw=gray!60, fill=gray!5, line width=1pt, dashed, inner sep=8pt, align=center},
  casenode/.style={rectangle, rounded corners=4pt, font=\small, align=center, inner sep=5pt, minimum width=4.5cm}
]

\node[phaselbl, text=blue!80!black] at (3.2, 8.3) {Symbiogenesis};
\node[font=\small\itshape, text=gray!60] at (3.2, 7.8) {Components fuse into composite systems};

\node[rectangle, rounded corners=4pt, draw=blue!70!black, fill=blue!12, text=blue!90!black, font=\small\bfseries, minimum width=1.6cm, minimum height=0.75cm, align=center] (llm) at (1.0, 6.5) {LLM};
\node[rectangle, rounded corners=4pt, draw=teal!70!black, fill=teal!12, text=teal!90!black, font=\small\bfseries, minimum width=1.6cm, minimum height=0.75cm, align=center] (code) at (3.0, 7.1) {Code\\Interpreter};
\node[rectangle, rounded corners=4pt, draw=purple!70!black, fill=purple!12, text=purple!90!black, font=\small\bfseries, minimum width=1.6cm, minimum height=0.75cm, align=center] (search) at (5.4, 6.5) {Search};

\node[rectangle, rounded corners=6pt, draw=blue!80!black, fill=blue!8, line width=1.5pt, inner sep=8pt, align=center] (agent) at (3.2, 3.5) {\Large Composite\\[-2pt]\Large Agentic System};

\draw[-{stealth}, thick, blue!70!black, line width=1.2pt] (llm.south) -- (agent.north west);
\draw[-{stealth}, thick, purple!70!black, line width=1.2pt] (search.south) -- (agent.north east);

\node[rectangle, rounded corners=4pt, draw=orange!70!black, fill=orange!12, text=orange!90!black, font=\small\bfseries, minimum width=1.6cm, minimum height=0.75cm, align=center] (memory) at (1.8, 5.3) {Memory};
\node[rectangle, rounded corners=4pt, draw=green!50!black, fill=green!10, text=green!70!black, font=\small\bfseries, minimum width=1.6cm, minimum height=0.75cm, align=center] (tools) at (4.6, 5.3) {Tool Use};

\draw[-{stealth}, thick, teal!70!black, line width=1.2pt] (code.south) -- (agent.north);
\draw[-{stealth}, thick, orange!70!black, line width=1.2pt] (memory.south) -- (agent.north west);
\draw[-{stealth}, thick, green!50!black, line width=1.2pt] (tools.south) -- (agent.north east);

\draw[-{stealth}, ultra thick, color=transitioncol, line width=2.5pt] (6.8, 3.5) -- (8.2, 3.5);
\node[font=\normalsize\bfseries\itshape, text=transitioncol, align=center] at (7.5, 2.8) {Early\\2026};

\node[phaselbl, text=transitioncol] at (12.2, 8.3) {Niche Construction};
\node[font=\small\itshape, text=gray!60] at (12.2, 7.8) {Composite systems modify their environments};

\node[rectangle, rounded corners=6pt, draw=transitioncol, fill=orange!5, line width=1.5pt, inner sep=8pt, align=center] (agent2) at (12.2, 3.5) {\Large Composite\\[-2pt]\Large Agentic System};

\node[envnode] (env1) at (9.0, 6.5) {\small Codebases and\\Repositories};
\node[envnode] (env2) at (12.2, 6.8) {\small Training\\Infrastructure};
\node[envnode] (env3) at (15.4, 6.5) {\small Physical\\World};

\draw[-{stealth}, thick, transitioncol, line width=1.2pt] (agent2.north west) -- (env1.south east);
\draw[-{stealth}, thick, transitioncol, line width=1.2pt] (agent2.north) -- (env2.south);
\draw[-{stealth}, thick, transitioncol, line width=1.2pt] (agent2.north east) -- (env3.south west);

\draw[-{stealth}, thick, red!60!black, line width=1.2pt, dashed] (env1.south) to[bend right=15] (agent2.west);
\draw[-{stealth}, thick, red!60!black, line width=1.2pt, dashed] (env2.west) to[bend right=20] (agent2.north west);
\draw[-{stealth}, thick, red!60!black, line width=1.2pt, dashed] (env3.south) to[bend left=15] (agent2.east);

\node[font=\scriptsize\itshape, text=red!60!black, align=center, fill=white, fill opacity=0.85, text opacity=1, inner sep=2pt, rounded corners=1pt] at (9.8, 4.6) {Modified environment\\reshapes selection\\pressures};

\draw[gray!30, line width=0.5pt] (0, 1.8) -- (16.5, 1.8);
\node[font=\small\bfseries, text=gray!50] at (8.25, 1.2) {Early 2026 Instantiations};

\node[casenode, draw=blue!50, fill=blue!5] at (2.5, -0.2) {\textbf{Claude Code}\\Orchestrated model +\\tools + sub-agents\\reshaping codebases};
\node[casenode, draw=transitioncol, fill=orange!5] at (8.25, -0.2) {\textbf{OpenClaw + Robotics}\\Agent skills deployed to\\physical hardware; robots\\expose capabilities back};
\node[casenode, draw=green!40!black, fill=green!5] at (14.0, -0.2) {\textbf{Karpathy / nanochat}\\AI agents optimize\\training infrastructure\\for next-gen models};

\end{tikzpicture}
}
\caption{From Symbiogenesis to Niche Construction. \textit{Left:} The Symbiogenesis pattern (2024--2025)---independent components (LLMs, code interpreters, search, memory, tool-use) fuse into composite agentic systems. \textit{Right:} The Niche Construction pattern (early 2026)---composite systems actively modify their deployment environments (codebases, training infrastructure, physical world), creating feedback loops where the modified environment reshapes the selection pressures on subsequent system generations. Dashed red arrows denote the feedback channel. Three early 2026 case studies instantiate this dynamic.}
\label{fig:nicheconstruction}
\end{figure}

This niche-constructive dynamic does not, in our assessment, constitute a phase transition in the formal sense of Definition~3 (Equation~\ref{eq:phase}). It has not caused the sudden redistribution of ecosystem configurations---the spike in $dH/dt$---that characterizes punctuation events like the DeepSeek Moment. Rather, it represents an intensification of Symbiogenesis dynamics: composite systems becoming increasingly integrated with, and increasingly capable of modifying, their deployment environments. The boundary between this intensification and the onset of Noogenesis (Section~6.1)---which would require closed-loop recursive self-improvement, where systems modify their own architectures and objectives rather than merely their operating environments---has not yet been crossed. But the distance to that boundary appears to be narrowing.

\subsection{The Frontier Lab Ecosystem: A Global Taxonomy}

\begin{figure}[H]
\centering
\resizebox{\textwidth}{!}{%
\begin{tikzpicture}[font=\sffamily, x=1.0cm, y=0.72cm]

\definecolor{usacol}{HTML}{3B6CB5}
\definecolor{chinacol}{HTML}{C44E52}
\definecolor{eurocol}{HTML}{E8963E}
\definecolor{othercol}{HTML}{6B8E6B}



\draw[gray!40, line width=0.5pt] (1.5, 8.65) -- (18.8, 8.65);
\draw[gray!40, line width=0.5pt] (1.5, 3.65) -- (18.8, 3.65);
\draw[gray!40, line width=0.5pt] (1.5, -0.65) -- (18.8, -0.65);

\node[font=\normalsize\bfseries, text=usacol, rotate=90, anchor=south] at (1.0, 11.5) {United States};
\node[font=\normalsize\bfseries, text=chinacol, rotate=90, anchor=south] at (1.0, 6.0) {China};
\node[font=\normalsize\bfseries, text=eurocol, rotate=90, anchor=south] at (1.0, 0.65) {Europe};
\node[font=\normalsize\bfseries, text=othercol, rotate=90, anchor=south] at (1.0, -2.5) {Other};

\fill[usacol, rounded corners=3pt] (2, 13.675) rectangle (18.5, 14.325);
\node[font=\small\bfseries, text=white] at (10.25, 14) {Google DeepMind (2010)};

\fill[usacol!85, rounded corners=3pt] (7, 12.675) rectangle (18.5, 13.325);
\node[font=\small\bfseries, text=white] at (12.75, 13) {OpenAI (2015)};

\fill[usacol!70, rounded corners=3pt] (5, 11.675) rectangle (18.5, 12.325);
\node[font=\small\bfseries, text=white] at (11.75, 12) {Meta AI / FAIR (2013)};

\fill[usacol!55, rounded corners=3pt] (13, 10.675) rectangle (18.5, 11.325);
\node[font=\small\bfseries, text=white] at (15.75, 11) {Anthropic (2021)};

\fill[usacol!40, rounded corners=3pt] (15, 9.675) rectangle (18.5, 10.325);
\node[font=\small\bfseries, text=usacol!80!black, anchor=east] at (14.8, 10) {xAI (2023)};

\fill[chinacol, rounded corners=3pt] (2, 7.675) rectangle (18.5, 8.325);
\node[font=\small\bfseries, text=white] at (10.25, 8) {Baidu / ERNIE (2010)};

\fill[chinacol!80, rounded corners=3pt] (9, 6.675) rectangle (18.5, 7.325);
\node[font=\small\bfseries, text=white] at (13.75, 7) {Alibaba / Qwen (2017)};

\fill[chinacol!60, rounded corners=3pt] (15, 5.675) rectangle (18.5, 6.325);
\node[font=\small\bfseries, text=chinacol!80!black, anchor=east] at (14.8, 6) {DeepSeek (2023)};

\fill[chinacol!45, rounded corners=3pt] (15, 4.675) rectangle (18.5, 5.325);
\node[font=\small\bfseries, text=chinacol!80!black, anchor=east] at (14.8, 5) {01.AI / Yi (2023)};

\fill[eurocol, rounded corners=3pt] (12, 2.675) rectangle (18.5, 3.325);
\node[font=\small\bfseries, text=white] at (15.25, 3) {Stability AI (UK, 2020)};

\fill[eurocol!75, rounded corners=3pt] (15, 1.675) rectangle (18.5, 2.325);
\node[font=\small\bfseries, text=eurocol!80!black, anchor=east] at (14.8, 2) {Mistral AI (France, 2023)};

\fill[eurocol!55, rounded corners=3pt] (16, 0.675) rectangle (18.5, 1.325);
\node[font=\small\bfseries, text=eurocol!80!black, anchor=east] at (15.8, 1) {Black Forest Labs (Germany, 2024)};

\fill[eurocol!40, rounded corners=3pt] (18, -0.325) rectangle (18.5, 0.325);
\node[font=\small\bfseries, text=eurocol!80!black, anchor=east] at (17.8, 0) {AMI Labs (France, 2026)};

\fill[othercol, rounded corners=3pt] (11, -2.325) rectangle (18.5, -1.675);
\node[font=\small\bfseries, text=white] at (14.75, -2) {Cohere (Canada, 2019)};

\fill[othercol!70, rounded corners=3pt] (9, -3.325) rectangle (18.5, -2.675);
\node[font=\small\bfseries, text=white] at (13.75, -3) {AI21 Labs (Israel, 2017)};

\draw[thick, gray!60] (1.5, -4.2) -- (18.8, -4.2);
\foreach \x/\yr in {2/2010, 4/2012, 6/2014, 8/2016, 10/2018, 12/2020, 14/2022, 16/2024, 18/2026} {
  \draw[gray!60, thick] (\x, -4.2) -- (\x, -4.45);
  \node[font=\small, text=gray!70!black, below] at (\x, -4.45) {\yr};
}

\foreach \x in {2, 4, 6, 8, 10, 12, 14, 16, 18} {
  \draw[gray!12, line width=0.4pt] (\x, -4.2) -- (\x, 14.5);
}

\draw[red!60!black, dashed, line width=0.8pt] (14.9, -4.2) -- (14.9, 15.0);
\node[fill=red!70!black, text=white, font=\scriptsize\bfseries, rounded corners=3pt, inner sep=3pt, align=center] at (14.9, 15.4) {ChatGPT Launch\\(Nov 2022)};

\draw[red!60!black, dashed, line width=0.8pt] (17, -4.2) -- (17, 15.0);
\node[fill=red!70!black, text=white, font=\scriptsize\bfseries, rounded corners=3pt, inner sep=3pt, align=center] at (17, 15.4) {DeepSeek R1\\(Jan 2025)};

\end{tikzpicture}
}
\caption{Global Frontier AI Labs---Founding Timeline, Country of Origin, and Active Period. Labs are grouped by region and sorted by founding date, with consistent color coding by geography. Vertical dashed lines mark the two defining punctuation events of the Generative AI era. The clustering of new lab formations in 2023 reflects the adaptive radiation dynamics of Epoch~II.}
\label{fig:labs}
\end{figure}
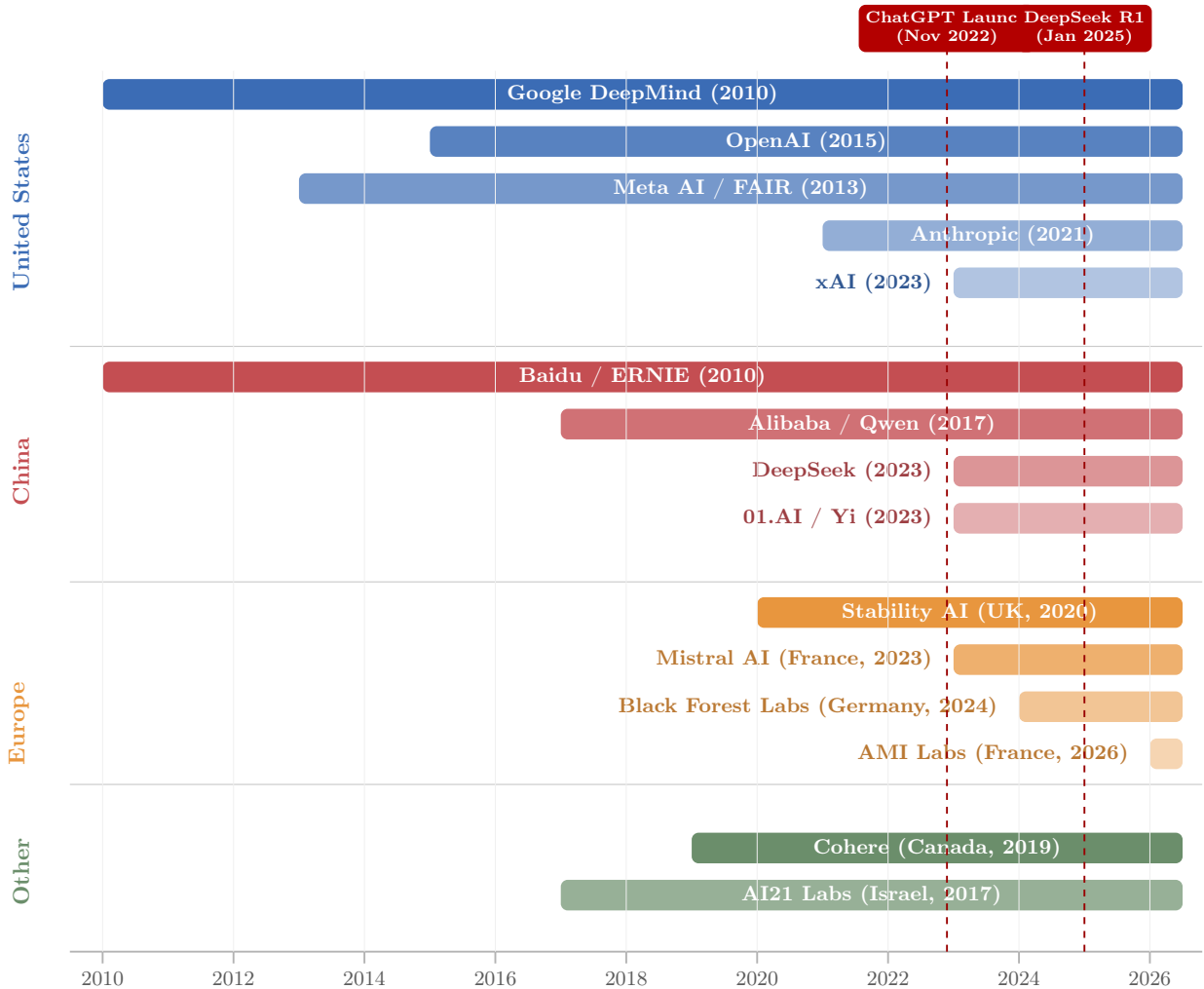

The Generative AI Era has produced a globally distributed ecosystem of frontier AI laboratories whose competitive dynamics drive the evolutionary process. These labs function as the `organisms' in our evolutionary framework: they compete for resources (capital, compute, talent, data), occupy ecological niches (consumer vs.\ enterprise, open vs.\ closed, general vs.\ specialized), and exert selection pressures on each other through benchmark competition and market positioning (see Figure~\ref{fig:labs} and Table~\ref{tab:labs}).

\textbf{United States.} OpenAI (founded 2015) pioneered the GPT series and the RLHF alignment paradigm; ChatGPT's November 2022 launch catalyzed the Great Expansion. Google DeepMind (formed April 2023 by merging Google Brain and DeepMind, est.\ 2010) inherits the combined legacy of its predecessor labs: Google Brain co-originated the Transformer \cite{vaswani2017} and produced BERT \cite{devlin2018} and PaLM \cite{chowdhery2022}; DeepMind developed AlphaFold \cite{jumper2021} and AlphaGo. The merged entity launched the Gemini series (December 2023). Anthropic (founded 2021 by former OpenAI researchers) developed Constitutional AI (RLAIF) and the Claude series through Claude Opus 4.6 (2026), emphasizing interpretability and safety. Meta AI/FAIR (Llama series from February 2023) led the open-weight movement with Llama 1/2/3/4. xAI (founded 2023 by Elon Musk) developed the Grok series and the Colossus 100,000-GPU training cluster.

\textbf{China.} DeepSeek (founded 2023, backed by quantitative trading firm High-Flyer) released DeepSeek Coder, V2, V3, and the landmark R1 reasoning model. Alibaba's Qwen team (from August 2023) produced the Qwen series through Qwen 3, with strong multilingual performance. Baidu (ERNIE series), Moonshot AI (Kimi K2 Thinking), MiniMax (M2), and 01.AI (Yi series) completed a deep competitive ecosystem. By late 2025, Chinese open-source models from Moonshot (Kimi K2) and MiniMax (M2) had begun outperforming select Western frontier models on specific benchmarks.

\textbf{Europe.} Mistral AI (founded 2023, Paris) emerged as Europe's leading frontier lab with Mixtral MoE and an Apache 2.0 release strategy emphasizing European sovereignty. Black Forest Labs (founded 2024, Germany) produced FLUX.1 for frontier image generation. Stability AI (founded 2019, UK) pioneered open-source image generation. AMI Labs (founded 2026, Paris), launched by Yann LeCun with a \texteuro890M seed round backed by NVIDIA and Temasek, is developing ``world model'' AI systems that learn from reality rather than language alone---representing a fundamental architectural challenge to the LLM-dominated paradigm.

\textbf{Other Markets.} Cohere (founded 2019, Canada) pursued an enterprise-first strategy with Command R. AI21 Labs (founded 2017, Israel) developed Jamba, a hybrid SSM-Transformer architecture.

\begin{table}[H]
\centering
\caption{Major Frontier AI Laboratories (as of March 2026)}
\label{tab:labs}
\small
\begin{tabularx}{\textwidth}{@{} >{\raggedright\arraybackslash}p{2.6cm} >{\raggedright\arraybackslash}p{1.4cm} >{\raggedright\arraybackslash}p{1.4cm} >{\raggedright\arraybackslash}p{2.2cm} >{\raggedright\arraybackslash}X @{}}
\toprule
\textbf{Laboratory} & \textbf{Country} & \textbf{Entry} & \textbf{First Model} & \textbf{Key Contributions} \\
\midrule
OpenAI & USA & Jun 2018 & GPT-1 (117M) & GPT series through GPT-5, o-series reasoning, DALL$\cdot$E, Codex, Stargate (\$500B) \\
Google DeepMind$^*$ & UK/USA & Dec 2023 & Gemini 1.0 & Predecessor labs co-originated the Transformer (2017) and BERT (2018); PaLM, Gemini series, AlphaFold \\
Anthropic & USA & Mar 2023 & Claude 1.0 & Constitutional AI (RLAIF), Claude through Opus 4.6 \\
Meta AI (FAIR) & USA & Feb 2023 & LLaMA (65B) & Llama open-weight series, PyTorch, open-source leadership \\
Mistral AI & France & Sep 2023 & Mistral 7B & Mixtral MoE, European sovereignty, Apache 2.0 \\
DeepSeek & China & Nov 2023 & DeepSeek Coder & R1 reasoning (Jan 2025), V3 MoE, cost-efficiency breakthroughs \\
Alibaba (Qwen) & China & Aug 2023 & Qwen-7B & Qwen series through Qwen 3, multilingual, open-weight \\
xAI & USA & Nov 2023 & Grok-1 & Grok series, Colossus 100K-GPU cluster \\
Moonshot AI & China & Oct 2023 & Kimi Chat & Kimi K2 Thinking, outperforming GPT-5 on select benchmarks \\
MiniMax & China & Dec 2023 & abab5.5 & M2 record open-model scores (2025) \\
Cohere & Canada & Nov 2022 & Command & Enterprise-first RAG, Command R series \\
AI21 Labs & Israel & Aug 2021 & Jurassic-1 & Jamba hybrid SSM-Transformer architecture \\
Stability AI & UK & Aug 2022 & Stable Diff.\ 1.0 & Open-source image generation \\
Black Forest Labs & Germany & Aug 2024 & FLUX.1 & Frontier image generation \\
AMI Labs & France & Mar 2026 & --- & World models (non-LLM), \texteuro890M seed, founded by Yann LeCun \\
\bottomrule
\end{tabularx}

\vspace{2pt}
{\footnotesize $^*$Formed April 2023 by merger of Google Brain and DeepMind (est.\ 2010). ``Entry'' date reflects the merged entity's first model release; seminal predecessor contributions---including the Transformer architecture \cite{vaswani2017} (Google Brain, 2017) and BERT \cite{devlin2018} (Google Brain, 2018)---predate the merger.}
\end{table}

The pace of frontier lab creation is itself accelerating: while OpenAI (2015) and DeepMind (2010) preceded the generative AI era, a cluster of new entrants---Anthropic, Mistral, xAI, DeepSeek, 01.AI, and Moonshot---all formed between 2021 and 2023, reflecting the adaptive radiation dynamics described in Section~4.2. The ecosystem now exhibits the classic dynamics of competitive exclusion and niche partitioning: labs differentiate on openness (Meta, Mistral vs.\ OpenAI, Anthropic), modality specialization (Black Forest Labs, Stability AI for images vs.\ OpenAI, Anthropic for text), geographic market focus (Qwen, DeepSeek for Chinese users vs.\ Cohere for enterprise), and capability tier (frontier reasoning vs.\ efficient deployment).

\subsubsection{The DeepSeek Moment: A Punctuation Event}

On January 20, 2025, Chinese AI laboratory DeepSeek released DeepSeek-R1 \cite{deepseek2025}, a reasoning model that matched OpenAI's o1 on multiple benchmarks at a reportedly under \$6 million training cost---using export-controlled H800 GPUs (the hobbled versions of NVIDIA's H100 approved for China under US sanctions). The market impact was immediate and severe: NVIDIA lost approximately \$589 billion in market capitalization on January 27---the largest single-day value loss in stock market history. The DeepSeek Moment constituted a punctuation event in our evolutionary framework. In the language of Definition~3, it caused a sudden spike in ecosystem entropy $dH/dt \gg \lambda_{\text{crit}}$: the prevailing assumption---that frontier AI required massive, US-controlled compute infrastructure---was rendered invalid within a single news cycle.

The event had three primary evolutionary consequences. \textit{Algorithmic Innovation as Compute Substitute.} DeepSeek's multi-head latent attention (MLA), mixture-of-experts routing, FP8 mixed-precision training, and GRPO-based reinforcement learning demonstrated that algorithmic efficiency could substitute for raw compute---compressing what had been considered a \$100M+ training run into a single-digit million-dollar operation. \textit{Open-Source Release as Competitive Strategy.} R1's release under the MIT License---the most permissive open-source license in common use---forced direct comparison with closed-source Western models and validated the open-weight paradigm Meta had pioneered with Llama. \textit{Export Control Failure as Selective Pressure.} US export restrictions designed to slow Chinese AI development had instead functioned as a selection pressure that forced efficiency innovations which proved transferable advantages. As RAND (Wang \& Siler-Evans, 2026) later reported, the operating costs of Chinese models now range between one-sixth and one-tenth of comparable US systems---a direct consequence of the efficiency innovations driven by hardware constraints.

\subsubsection{The Lunar New Year Effect}

By 2026, the entire Chinese AI ecosystem had adapted to the DeepSeek precedent, producing a coordinated wave of frontier releases designed to capture maximum attention during China's peak digital engagement period. In the six weeks preceding this paper: Alibaba released Qwen 3.5 (February 16, 2026); ByteDance launched Doubao 2.0; Zhipu AI unveiled GLM-5, trained entirely on Huawei Ascend chips (further demonstrating independence from NVIDIA); and multiple Western releases followed in quick succession (Claude Opus 4.6, GPT-5.3 Codex, Gemini 3.1 Pro). The pattern has become periodic: the ecosystem now generates synchronized bursts of frontier releases---consistent with punctuated equilibrium dynamics---rather than steady incremental progress. The Lunar New Year release wave functions as a cultural Schelling point: labs coordinate implicitly, knowing competitors will release during the same window, creating a self-reinforcing cycle of concentrated innovation.

\subsection{The Accelerating Evolution of Post-Training Alignment Methods}

\begin{figure}[H]
\centering
\resizebox{\textwidth}{!}{%
\begin{tikzpicture}[
  font=\sffamily,
  x=4.0cm, y=1cm,
  dominant/.style={rectangle, rounded corners=5pt, minimum width=2.8cm, minimum height=1.0cm,
    draw=#1!60!black, line width=1pt, fill=#1!12, font=\small\bfseries, align=center, text=black!85},
  variant/.style={rectangle, rounded corners=3pt, minimum width=2.2cm, minimum height=0.7cm,
    draw=#1!40!black, line width=0.6pt, fill=#1!8, font=\scriptsize\bfseries, align=center, text=black!80},
  arr/.style={-{stealth}, thick, gray!50}
]

\fill[red!5] (-0.3, -2.6) rectangle (2.15, 6.8);
\fill[blue!5] (2.15, -2.6) rectangle (4.5, 6.8);
\draw[gray!30, dashed, line width=0.8pt] (2.15, -2.6) -- (2.15, 6.8);
\node[font=\normalsize\bfseries\itshape, text=red!50!black] at (0.93, 6.4) {Epoch III: Great Expansion};
\node[font=\normalsize\bfseries\itshape, text=blue!50!black] at (3.35, 6.4) {Epoch IV: Symbiogenesis};

\draw[thick, gray!60] (-0.15, -2.0) -- (4.35, -2.0);
\foreach \yr/\x in {2022/0, 2023/1, 2024/2, 2025/3, 2026/4} {
  \draw[gray!60, thick] (\x, -2.0) -- (\x, -2.2);
  \node[font=\normalsize\bfseries, text=gray!70!black, below] at (\x, -2.2) {\yr};
}

\draw[decorate, decoration={brace, amplitude=4pt, mirror}, thick, gray!50]
    (-0.1, -3.0) -- (0.9, -3.0)
    node[midway, below=5pt, font=\scriptsize\bfseries, text=gray!60] {3 models};
\draw[decorate, decoration={brace, amplitude=4pt, mirror}, thick, gray!50]
    (1.0, -3.0) -- (2.1, -3.0)
    node[midway, below=5pt, font=\scriptsize\bfseries, text=gray!60] {2 models};
\draw[decorate, decoration={brace, amplitude=4pt, mirror}, thick, gray!50]
    (2.2, -3.0) -- (4.3, -3.0)
    node[midway, below=5pt, font=\scriptsize\bfseries, text=gray!60] {1 model};

\node[dominant=blue] (rlhf) at (0.4, 4.5) {RLHF + PPO\\[-2pt]{\scriptsize InstructGPT}};
\node[dominant=blue] (dpo) at (1.5, 4.5) {DPO\\[-2pt]{\scriptsize Direct Preference}};
\node[dominant=orange] (simpo) at (2.35, 4.5) {SimPO / ORPO\\[-2pt]{\scriptsize Reference-Free}};
\node[dominant=teal] (grpo) at (3.2, 4.5) {GRPO\\[-2pt]{\scriptsize DeepSeek R1}};
\node[dominant=purple] (cape) at (4.05, 4.5) {CAPE\\[-2pt]{\scriptsize Specification}};

\node[variant=gray] (sft) at (0.0, 1.5) {SFT\\[-1pt]{\tiny Supervised Fine-Tuning}};
\node[variant=purple] (cai) at (0.9, 1.5) {Constitutional AI\\[-1pt]{\tiny RLAIF}};
\node[variant=teal] (ipo) at (1.85, 1.5) {IPO / KTO\\[-1pt]{\tiny Variants}};
\node[variant=teal!70!blue] (rlvr) at (2.75, 1.5) {RLVR\\[-1pt]{\tiny Verifiable Rewards}};
\node[variant=purple] (dapo) at (3.65, 1.5) {DAPO / Dr.GRPO\\[-1pt]{\tiny Variants}};

\draw[arr] (rlhf) -- (dpo);
\draw[arr] (dpo) -- (simpo);
\draw[arr] (simpo) -- (grpo);
\draw[arr] (grpo) -- (cape);

\draw[arr, gray!35] (sft) -- (rlhf);
\draw[arr, gray!35] (cai) -- (dpo);
\draw[arr, gray!35] (ipo) -- (simpo);
\draw[arr, gray!35] (rlvr) -- (grpo);
\draw[arr, gray!35] (dapo) -- (cape);

\node[font=\scriptsize\itshape, text=blue!60!black] at (0.4, 3.6) {$\sim$18 months};
\node[font=\scriptsize\itshape, text=blue!60!black] at (1.5, 3.6) {$\sim$12 months};
\node[font=\scriptsize\itshape, text=orange!60!black] at (2.35, 3.6) {$\sim$6 months};
\node[font=\scriptsize\itshape, text=teal!60!black] at (3.2, 3.6) {$\sim$weeks};

\node[font=\small\bfseries, text=gray!60, rotate=90, anchor=south] at (-0.55, 4.5) {Dominant};
\node[font=\small\bfseries, text=gray!60, rotate=90, anchor=south] at (-0.55, 1.5) {Variants};

\end{tikzpicture}
}
\caption{Evolution of Post-Training Alignment Methods---From RLHF to GRPO and Beyond (2022--2026). Dominant paradigms (top) are superseded at accelerating rates ($\sim$18 months $\to$ $\sim$12 months $\to$ weeks), while variant methods (bottom) feed innovations into the next dominant paradigm. The pipeline simplifies progressively from three models to one.}
\label{fig:alignment}
\end{figure}
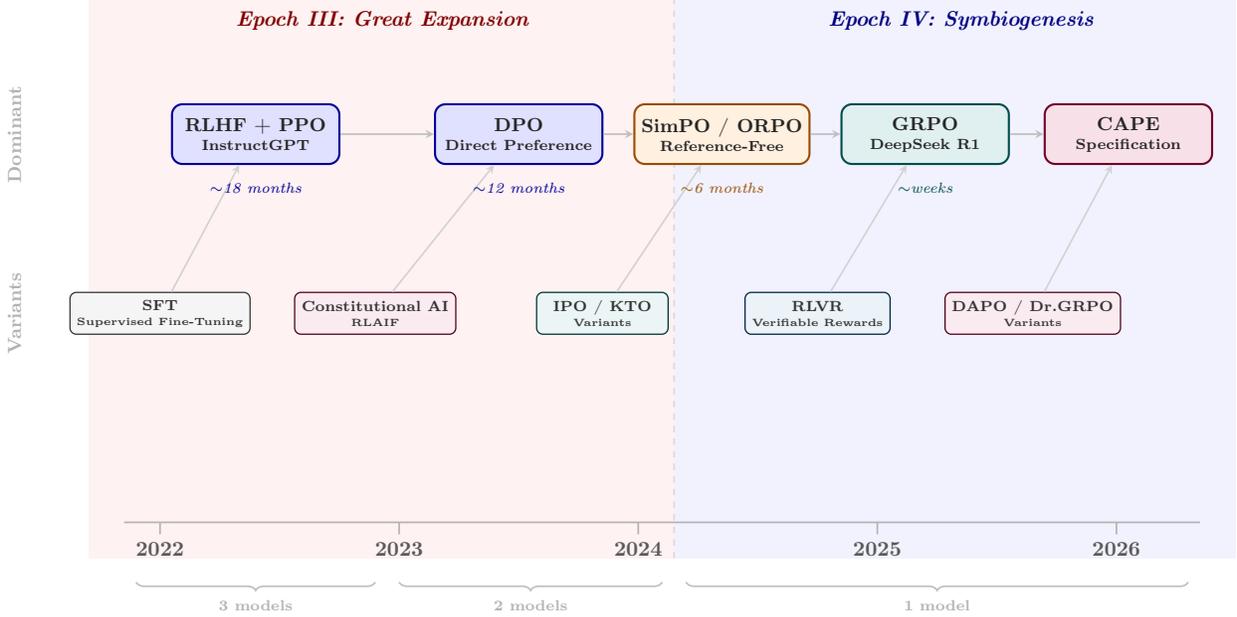

Post-training alignment---the process of refining a pre-trained language model to follow human instructions, produce helpful and truthful outputs, and avoid harmful behavior---has itself undergone a rapid evolutionary sequence. The paradigm turnover rate is itself accelerating, with each dominant method being superseded in progressively shorter intervals (see Figure~\ref{fig:alignment}).

\textbf{Phase 1: RLHF + PPO (2022--2023, $\sim$18 months dominant).} The original alignment paradigm, documented by Ouyang et al.\ \cite{ouyang2022} and powered by Proximal Policy Optimization \cite{schulman2017}, required a three-model pipeline: a policy model (the language model being aligned), a reward model (trained on human preference data to predict which outputs humans prefer), and a value/critic model (estimating expected cumulative reward). The training objective:
\begin{equation}
\label{eq:rlhf}
\pi^* = \argmax_\pi \mathbb{E}_{x,y\sim\pi}\bigl[r_\varphi(x,y)\bigr] - \beta \cdot D_{\text{KL}}(\pi \| \pi_{\text{ref}})
\end{equation}
where the KL-divergence term prevents the aligned model from deviating too far from the pre-trained base. This enabled ChatGPT and powered the first wave of instruction-following models. Limitations: expensive data collection (requiring expert human annotators), training instability, and reward model exploitation (``reward hacking'').

\textbf{Phase 2: Constitutional AI / RLAIF (2022--2023).} Anthropic's Constitutional AI \cite{bai2022} replaced human labelers with AI-generated critique, guided by explicit constitutional principles. This was an evolutionary efficiency gain: reducing the cost of the feedback signal while maintaining---and in some cases improving---alignment quality. RLAIF (RL from AI Feedback) shifted the bottleneck from human labor to principle engineering, foreshadowing the specification-based methods that would emerge later.

\textbf{Phase 3: DPO (2023--2024, $\sim$12 months dominant).} Direct Preference Optimization \cite{rafailov2023} represented a significant architectural simplification. By reparameterizing the RLHF objective, Rafailov et al.\ showed that the optimal policy could be extracted directly from preference data without training a separate reward model:
\begin{equation}
\label{eq:dpo}
\mathcal{L}_{\text{DPO}} = -\mathbb{E}\!\left[\log \sigma\!\left(\beta \log \frac{\pi_\theta(y_w|x)}{\pi_{\text{ref}}(y_w|x)} - \beta \log \frac{\pi_\theta(y_l|x)}{\pi_{\text{ref}}(y_l|x)}\right)\right]
\end{equation}
This reduced the three-model pipeline to two models (policy + reference), dramatically simplifying training infrastructure. DPO became the dominant alignment method for open-source models during 2023--2024. Variants proliferated: IPO, KTO, SimPO---each optimizing different aspects of the preference learning problem, demonstrating the rapid speciation characteristic of adaptive radiation.

\textbf{Phase 4: GRPO / RLVR (2024--present, current dominant paradigm).} Group Relative Policy Optimization \cite{shao2024,deepseek2025} eliminated both the reward model and the critic network. For each prompt $x$, GRPO samples $G$ responses, scores them using a verifiable reward function (e.g., mathematical correctness, code execution success), and normalizes advantages within the batch:
\begin{equation}
\label{eq:grpo}
\mathcal{L}_{\text{GRPO}} = -\frac{1}{G}\sum_{i=1}^{G}\min\!\left(\frac{\pi_\theta(o_i|x)}{\pi_{\text{old}}(o_i|x)}\hat{A}_i,\; \text{clip}\!\left(\frac{\pi_\theta}{\pi_{\text{old}}}, 1\!\pm\!\epsilon\right)\!\hat{A}_i\right)
\end{equation}
This produces a single-model training pipeline: only the policy model is needed. GRPO's reliance on verifiable rewards (which can be checked algorithmically rather than requiring human judgment) proved decisive in DeepSeek-R1's training, enabling the model to develop sophisticated chain-of-thought reasoning through pure reinforcement learning without any supervised fine-tuning on reasoning traces. The paradigm turnover rate is itself accelerating: RLHF dominated for approximately 18 months, DPO for approximately 12 months, and GRPO variants now evolve on timescales of weeks.

The evolutionary trajectory of alignment methods reveals a clear trend toward simplification: from three-model RLHF to two-model DPO to single-model GRPO, each step reducing infrastructure complexity while maintaining or improving alignment quality. This mirrors biological evolution's tendency toward metabolic efficiency---organisms that achieve the same function with less energy expenditure hold an adaptive advantage. The question of what comes after GRPO---potentially specification-based alignment, where models are aligned through declarative behavioral specifications rather than any form of preference learning---represents a potential Phase 5 that would constitute yet another paradigm shift.

Han et al.\ \cite{han2025} demonstrated the quantization trap---compression paradoxically increases energy while degrading multi-hop reasoning:
\begin{equation}
\label{eq:qtrap}
E_q(b, d) \propto b^{-1} \cdot d \cdot \gamma_{\text{grid}}, \quad \gamma_{\text{grid}} \gg 1 \;\text{for low}\; b
\end{equation}
This creates a structural tension directly relevant to alignment evolution: the multi-step reasoning capability that makes GRPO-trained models powerful (extended chain-of-thought with 10--50+ reasoning steps) is precisely the cognitive modality that quantization degrades most severely. A GRPO-aligned reasoning model quantized from 16-bit to 4-bit may lose the very capability that alignment was designed to elicit. This coupling between alignment method innovation and deployment constraints further supports the Institutional Scaling Law's prediction of environment-specific optima: the `right' alignment method depends on the deployment environment's precision, latency, and trust constraints.

\section{The Rise of Sovereign AI}

\begin{figure}[H]
\centering
\resizebox{\textwidth}{!}{%
\begin{tikzpicture}[font=\sffamily, x=1.0cm, y=1cm]

\definecolor{americol}{HTML}{3B6CB5}
\definecolor{asiacol}{HTML}{C44E52}
\definecolor{eurocol}{HTML}{E8963E}
\definecolor{mideastcol}{HTML}{2E8B7B}
\definecolor{othercol}{HTML}{6B8E6B}

\foreach \v in {2, 4, 6, 8, 10, 12} {
  \draw[gray!12, line width=0.4pt] (\v, -2.0) -- (\v, 11.5);
}

\draw[gray!40, line width=0.5pt] (0, 8.65) -- (12.5, 8.65);
\draw[gray!40, line width=0.5pt] (0, 4.65) -- (12.5, 4.65);
\draw[gray!40, line width=0.5pt] (0, 1.65) -- (12.5, 1.65);

\node[font=\scriptsize\bfseries, text=americol, rotate=90, anchor=south] at (-0.5, 9.5) {Americas};
\node[font=\scriptsize\bfseries, text=asiacol, rotate=90, anchor=south] at (-0.5, 6.5) {Asia-Pacific};
\node[font=\scriptsize\bfseries, text=mideastcol, rotate=90, anchor=south] at (-0.5, 3.0) {Middle East};
\node[font=\scriptsize\bfseries, text=eurocol, rotate=90, anchor=south] at (-0.5, 0.5) {Europe};

\fill[americol, rounded corners=3pt] (0, 9.675) rectangle (11, 10.325);
\node[font=\scriptsize\bfseries, text=white] at (5.5, 10) {USA (est.\ 2023)};

\fill[americol!45, rounded corners=3pt] (0, 8.675) rectangle (2, 9.325);
\node[font=\scriptsize\bfseries, text=americol!80!black, anchor=west] at (2.2, 9) {Latin America (est.\ 2025)};

\fill[asiacol, rounded corners=3pt] (0, 7.675) rectangle (9.5, 8.325);
\node[font=\scriptsize\bfseries, text=white] at (4.75, 8) {China (est.\ 2023)};

\fill[asiacol!70, rounded corners=3pt] (0, 6.675) rectangle (4.5, 7.325);
\node[font=\scriptsize\bfseries, text=white] at (2.25, 7) {South Korea (est.\ 2025)};

\fill[asiacol!55, rounded corners=3pt] (0, 5.675) rectangle (4, 6.325);
\node[font=\scriptsize\bfseries, text=white] at (2.0, 6) {India (est.\ 2024)};

\fill[asiacol!40, rounded corners=3pt] (0, 4.675) rectangle (3.5, 5.325);
\node[font=\scriptsize\bfseries, text=asiacol!80!black] at (1.75, 5) {Japan (est.\ 2024)};

\fill[mideastcol, rounded corners=3pt] (0, 3.675) rectangle (5.5, 4.325);
\node[font=\scriptsize\bfseries, text=white] at (2.75, 4) {UAE (est.\ 2024)};

\fill[mideastcol!70, rounded corners=3pt] (0, 2.675) rectangle (5, 3.325);
\node[font=\scriptsize\bfseries, text=white] at (2.5, 3) {Saudi Arabia (est.\ 2025)};

\fill[mideastcol!45, rounded corners=3pt] (0, 1.675) rectangle (2.5, 2.325);
\node[font=\scriptsize\bfseries, text=mideastcol!80!black, anchor=west] at (2.7, 2) {Switzerland (est.\ 2025)};

\fill[eurocol, rounded corners=3pt] (0, 0.675) rectangle (8.5, 1.325);
\node[font=\scriptsize\bfseries, text=white] at (4.25, 1) {EU / France (est.\ 2024)};

\fill[eurocol!70, rounded corners=3pt] (0, -0.325) rectangle (6, 0.325);
\node[font=\scriptsize\bfseries, text=white] at (3.0, 0) {UK (est.\ 2024)};

\draw[thick, gray!60] (0, -1.2) -- (12.5, -1.2);
\foreach \v in {0, 2, 4, 6, 8, 10, 12} {
  \draw[gray!60, thick] (\v, -1.2) -- (\v, -1.45);
  \node[font=\scriptsize, text=gray!70!black, below] at (\v, -1.45) {\v};
}
\node[font=\scriptsize\bfseries, text=gray!60] at (6.25, -2.2) {Relative Sovereign AI Investment \& Maturity Score (conceptual)};

\end{tikzpicture}
}
\caption{The Rise of Sovereign AI---National Programs and Relative Investment Scale (2023--2026). Major national sovereign AI programs grouped by region and ranked by investment scale, showing the global distribution of sovereign AI initiatives across six continents.}
\label{fig:sovereign}
\end{figure}
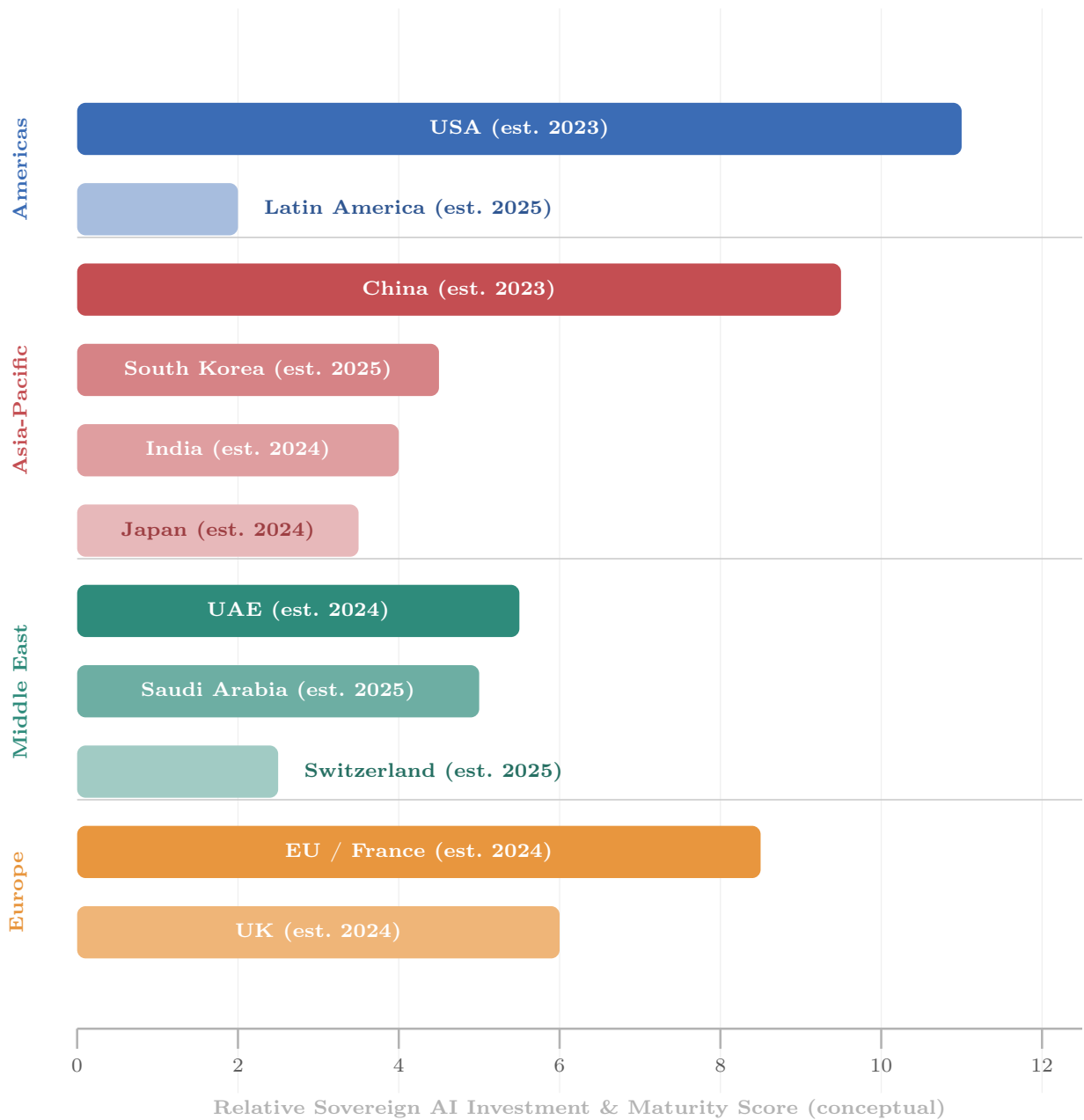

The concept of Sovereign AI---the assertion that nation-states and institutions must control the AI systems that influence their citizens, economies, and security---has emerged as the defining geopolitical phenomenon of the Symbiogenesis epoch. In our evolutionary framework (Section~3.1), sovereign AI represents a new class of environmental selection pressure that drives model speciation: as nations impose distinct requirements on data residency, linguistic performance, regulatory compliance, and cultural alignment, the optimal AI configuration $\theta^*(\varepsilon)$ diverges across environments, producing an ecosystem of jurisdictionally adapted models (Figure~\ref{fig:sovereign}). This section documents the empirical evidence for this speciation.

\subsection{Defining Sovereign AI}

Cellucci and Singh \cite{cellucci2025} define Sovereign AI as the principle that states and institutions must maintain control over four dimensions of the AI stack: (1) \textit{Training Data Sovereignty}---curating and controlling training corpora that reflect national languages, legal frameworks, cultural norms, and institutional knowledge; (2) \textit{Model Sovereignty}---developing, fine-tuning, or commissioning models architecturally designed for institutional missions; (3) \textit{Infrastructure Sovereignty}---maintaining compute, storage, and inference infrastructure under national or institutional jurisdiction; and (4) \textit{Interaction Sovereignty}---ensuring that prompts, queries, and model outputs remain within sovereign data boundaries. Each dimension corresponds to a component of our Sovereignty Compliance index $\Sigma(\theta, \varepsilon)$ in Definition~1.

McKinsey projects sovereign AI could represent a \$600 billion market by 2030. This projection treats sovereign AI as a market; our framework treats it as an evolutionary force---one that restructures the entire fitness landscape of AI deployment, driving speciation toward jurisdictionally adapted models and away from the universalist paradigm assumed by classical scaling laws.

\subsection{National Sovereign AI Programs: A Competitive Taxonomy}

\textbf{United States.} The Stargate Project (\$500 billion AI infrastructure initiative announced January 2025) represents the most ambitious sovereign compute investment to date. The Stargate Project is a collaboration between OpenAI, Oracle, SoftBank, and the US government to build AI data centers across Texas and other states, establishing US compute supremacy for the next generation of frontier model training. The Trump administration has adopted a deregulatory stance, positioning the US as the global leader in AI innovation while resisting binding international AI governance frameworks.

\textbf{China.} China's AI research output in 2024 matched the combined publications of the US, UK, and EU. The Chinese government has invested over \$140 billion in AI development. Huawei's Ascend chip ecosystem (910B, 920 series) and SMIC's manufacturing advances are reducing reliance on Western semiconductors. DeepSeek, Qwen, and a constellation of open-source Chinese models have demonstrated frontier performance at a fraction of Western development costs. China's AI governance framework combines strong state direction with aggressive industrial policy, producing an ecosystem that prioritizes capability, cost efficiency, and strategic self-sufficiency.

\textbf{European Union.} The EU has pursued a regulatory-first approach to AI sovereignty, anchored by the AI Act (2024). The OpenEuroLLM initiative is developing open LLMs across 24 official EU languages. AI Factories are being established based on EuroHPC Joint Undertaking supercomputers. France has secured \texteuro{}109 billion in AI investment commitments, with President Macron positioning France as Europe's AI hub. Mistral AI leads European sovereign LLM development with open-weight models released under Apache 2.0.

\textbf{Middle East.} The UAE, through G42 and the Mohamed bin Zayed University of AI (MBZUAI), has developed open-weight Arabic-centric LLMs (Falcon, Jais). Saudi Arabia launched HUMAIN, backed by its sovereign wealth fund, as a full-stack AI ecosystem encompassing compute infrastructure, model development, and application deployment. The Gulf states are positioning AI sovereignty as central to their post-oil economic diversification strategies.

\textbf{Asia-Pacific.} India's IndiaAI Mission (\textrm{INR} 10,300+ crore) encompasses sovereign compute, model development, and AI education. India hosted the first Global South AI summit (February 2026, see Section~5.4). Sarvam AI launched 30B/105B MoE models for India's multilingual landscape of 22 official languages. South Korea announced plans with NVIDIA to deploy 260,000+ GPUs across sovereign cloud infrastructure. Japan allocated \$1.5B for AI semiconductor development through its METI-led strategy.

\subsection{Davos 2026: Sovereign AI as Global Consensus}

The World Economic Forum's 56th Annual Meeting (January 19--23, 2026 in Davos-Klosters, Switzerland) convened nearly 3,000 leaders from 133 countries under the theme ``A Spirit of Dialogue.'' The technology agenda was dominated by AI, and specifically by sovereign AI. Several developments marked a qualitative shift from previous summits. WEF/Bain projected global AI investment to reach \$1.5 trillion for applications and \$400 billion for infrastructure annually by 2030 \cite{wefbain2026}. The conversation shifted from \textit{whether} to govern AI to \textit{how} to operationalize governance at the national and institutional level. Anthropic CEO Dario Amodei warned that AI would produce ``very high GDP growth and potentially also very high unemployment and inequality.'' Google DeepMind CEO Demis Hassabis stated that current AI systems remain ``nowhere near'' human-level AGI, tempering speculative forecasts. The summit revealed a near-universal consensus that sovereign AI is not a niche concern but a central element of national strategy. Even small nations---Estonia, Singapore, Rwanda---announced AI sovereignty initiatives scaled to their resources, confirming the Speciation prediction (Proposition~1): sufficiently distinct environments produce distinct AI optima.

\subsection{The India AI Impact Summit 2026}

India's AI Impact Summit (February 16--20, 2026 in New Delhi) was the first global AI summit hosted in the Global South under India's G20 presidency. The summit drew delegations from over 100 countries, including 20 heads of state. Investment commitments exceeded \$200 billion.

The summit's significance for our evolutionary framework lies in three dynamics. First, the White House rejected global AI governance outright, with envoy David Sacks stating: ``We do not think that global governance is necessary, or frankly desirable.'' This explicit rejection of AI multilateralism accelerates the speciation process---without a global governance framework, national AI ecosystems will diverge faster, producing the fragmented landscape predicted by Proposition~1. Second, middle powers (India, Brazil, UAE, Indonesia) increasingly seek to build independent AI capabilities rather than rely on US or Chinese models---precisely the selection pressure driving the adaptive radiation described in Section~4.2, now operating at the geopolitical rather than the technological level. Third, India's Union Minister Ashwini Vaishnaw outlined India's ``whole-of-nation'' AI strategy as one built on ``frugal, sovereign, and scalable'' AI---a direct expression of the Institutional Scaling Law's prediction that cost-constrained environments will converge on smaller, domain-specific optimal scales $N^*(\varepsilon) \ll N_{\text{frontier}}$.

\subsection{Sovereign AI as Evolutionary Selection Pressure}

From an evolutionary perspective, sovereign AI introduces a new class of \textit{ecological niche} that drives adaptive radiation at the geopolitical level. Just as geographic isolation produces speciation in biology, jurisdictional isolation---different data protection regimes, different linguistic requirements, different cultural norms, different strategic priorities---produces model specialization. Models must differentiate not only on capability benchmarks but on cultural attunement, linguistic coverage, regulatory compliance, and data provenance. This is the biological equivalent of \textit{character displacement}: when two species occupy the same geographic area but different ecological niches, they diverge in traits relevant to niche exploitation, reducing direct competition. The result is not one dominant global AI model but an ecosystem of culturally and jurisdictionally adapted models, connected through translation layers and interoperability standards but fundamentally distinct in their optimization targets. The Institutional Scaling Law (Equation~\ref{eq:inst_scaling}) provides the formal structure: the weight vector $\w(\varepsilon)$ varies across sovereign environments, producing different optimal scales $N^*(\varepsilon)$ and different fitness landscapes---and the Speciation Proposition (Equation~\ref{eq:speciation}) guarantees that sufficiently different environments produce divergent optima.

\begin{sloppypar}
The simultaneous democratization of training infrastructure (Section~\ref{sec:early2026}) interacts with these sovereign dynamics in a mutually reinforcing manner. Sovereign mandates create the \textit{demand} for jurisdictionally adapted models---the environmental heterogeneity that drives speciation. The reduction of training barriers---exemplified by projects like Karpathy's nanochat, which reproduces GPT-2 in two hours on commodity hardware---provides the \textit{supply-side} mechanism by which institutions can reach their local fitness optima. The combined effect accelerates the fragmentation that the Speciation Proposition describes: national programs that would otherwise depend on frontier labs for model access can increasingly train, fine-tune, and deploy at their environment-specific $N^*(\varepsilon)$. Notably, this ecological divergence occurs atop architectural convergence---most sovereign models remain transformers trained with GRPO variants---mirroring the biological pattern in which a shared body plan (the vertebrate skeleton, the transformer architecture) supports radical niche diversification.
\end{sloppypar}

\section{Forecasting Future Evolution}

The evolutionary framework and mathematical formalization developed in this paper can be used to forecast potential characteristics of future epochs and eras.

\subsection{The Next Epoch: Noogenesis (\texorpdfstring{$\sim$}{\textasciitilde}2026--2030?)}

The current Symbiogenesis epoch will likely yield to what we tentatively call `Noogenesis'---the emergence of genuinely autonomous cognitive systems capable of independent discovery and self-directed improvement. Key indicators include: autonomous scientific discovery (a 27B Gemma-based model with Sakana AI's AI Scientist framework generated a validated cancer hypothesis published in Nature Medicine in October 2025); self-improving architectures (GPT-5 Codex can work independently on complex coding projects for 7+ hours); sovereign AI ecosystem fragmentation (each major power bloc developing independent model lineages); API cost collapse (30$\times$ reduction in three years, Epoch AI data); and domain-specific model compression (2B--10B parameter models exceeding generalist performance on domain-specific tasks, while sidestepping the quantization trap by running at full precision on commodity hardware). This last point connects directly to the Institutional Scaling Law prediction: smaller, domain-specific models can achieve higher institutional fitness than frontier generalists precisely because they avoid the trust erosion ($\partial \T/\partial N < 0$) and cost penalties that suppress the fitness of larger models. Crucially, these smaller models avoid the reasoning accuracy degradation and paradoxical energy cost increases that plague compressed frontier deployments \cite{han2025}---achieving both higher task-specific trustworthiness and lower environmental cost per query.

The developments documented in Section~\ref{sec:early2026} bear directly on the proximity of this transition. Agentic systems that autonomously execute multi-hour coding sessions, training pipelines that optimize themselves through AI-contributed code changes, and agent frameworks that deploy capabilities to physical robots all satisfy necessary---but not sufficient---conditions for Noogenesis. The critical threshold remains closed-loop recursive self-improvement: systems that modify their own architectures, training procedures, and objective functions, not merely their operating environments. The niche-constructive dynamics observable in early 2026 (Section~\ref{sec:early2026}) narrow the distance to this boundary without crossing it. In particular, the AI-assisted training optimization pattern---where agents contribute code changes that reduce training time, which in turn accelerates the next cycle of model development---represents an open-loop approximation of recursive self-improvement that could, with sufficient integration, close into a genuine self-modifying cycle. Monitoring whether this loop closes is, in our assessment, the most informative leading indicator of the Symbiogenesis--Noogenesis boundary. Figure~\ref{fig:noogenesis} summarizes the current status of transition indicators.

\begin{figure}[H]
\centering
\resizebox{\textwidth}{!}{%
\begin{tikzpicture}[
  font=\sffamily,
  indicator/.style={rectangle, rounded corners=3pt, minimum width=13.5cm, minimum height=0.72cm, font=\small, inner sep=4pt},
  met/.style={indicator, draw=green!60!black, fill=green!8, text=black!80},
  partial/.style={indicator, draw=orange!70!black, fill=orange!8, text=black!80},
  notmet/.style={indicator, draw=red!50!black, fill=red!6, text=black!80},
  statusdot/.style={circle, inner sep=3.5pt, draw=#1!70!black, fill=#1!60, line width=0.8pt}
]

\node[font=\large\bfseries, text=black!80] at (7.5, 8.2) {Symbiogenesis $\to$ Noogenesis: Transition Indicator Assessment};
\node[font=\small\itshape, text=gray!60] at (7.5, 7.7) {Status as of March 2026};

\node[font=\normalsize\bfseries, text=green!50!black] at (7.5, 7.0) {Necessary conditions (met or substantially met)};

\node[statusdot=green!60!black] at (0.4, 6.2) {};
\node[met] at (7.5, 6.2) {Autonomous scientific discovery --- 27B model generated validated cancer hypothesis (Nature Medicine, Oct 2025)};

\node[statusdot=green!60!black] at (0.4, 5.35) {};
\node[met] at (7.5, 5.35) {Extended autonomous operation --- Claude Code 99.9th pctl.\ turns exceed 45 min; GPT-5 Codex operates 7+ hours};

\node[statusdot=green!60!black] at (0.4, 4.5) {};
\node[met] at (7.5, 4.5) {Sovereign AI ecosystem fragmentation --- 20+ national programs, divergent $\mathbf{w}(\varepsilon)$ across jurisdictions};

\node[statusdot=green!60!black] at (0.4, 3.65) {};
\node[met] at (7.5, 3.65) {API cost collapse and training democratization --- $30\times$ cost reduction; GPT-2 trainable in 2 hours on commodity GPUs};

\node[statusdot=green!60!black] at (0.4, 2.8) {};
\node[met] at (7.5, 2.8) {Niche construction by composite systems --- agentic systems modifying codebases, training infra, physical environments};

\node[font=\normalsize\bfseries, text=red!50!black] at (7.5, 1.9) {Sufficient conditions (not yet met)};

\node[statusdot=orange] at (0.4, 1.1) {};
\node[partial] at (7.5, 1.1) {AI-assisted training optimization --- agents contribute code changes to training pipelines (open-loop, not yet closed)};

\node[statusdot=red] at (0.4, 0.25) {};
\node[notmet] at (7.5, 0.25) {Closed-loop recursive self-improvement --- systems modifying own architectures, training procedures, and objectives};

\node[statusdot=red] at (0.4, -0.6) {};
\node[notmet] at (7.5, -0.6) {Autonomous objective setting --- systems defining their own goals independent of human specification};

\node[statusdot=green!60!black] at (4.0, -1.5) {};
\node[font=\small, anchor=west] at (4.4, -1.5) {Met};
\node[statusdot=orange] at (6.0, -1.5) {};
\node[font=\small, anchor=west] at (6.4, -1.5) {Partially met};
\node[statusdot=red] at (9.5, -1.5) {};
\node[font=\small, anchor=west] at (9.9, -1.5) {Not yet observed};

\end{tikzpicture}
}
\caption{Noogenesis Transition Indicators---Status Assessment (March 2026). Five necessary conditions for the Symbiogenesis--Noogenesis epoch transition are substantially met (green). The sufficient conditions---closed-loop recursive self-improvement and autonomous objective setting---remain unmet (red), with AI-assisted training optimization representing a partially met intermediate state (orange). The critical gap between necessary and sufficient conditions defines the current distance to the epoch boundary.}
\label{fig:noogenesis}
\end{figure}

\subsection{The Next Era: The Post-Transformer (?)}

A new era in our taxonomy requires a \textit{discontinuous substrate change}---a phase transition event that renders the current computational paradigm subordinate, just as the transformer rendered recurrent architectures subordinate in 2017. Candidates include: state-space models (SSMs), exemplified by Mamba, which process sequences in linear rather than quadratic time; neuromorphic computing, which replaces von Neumann architecture with brain-inspired spike-based processing; and quantum-enhanced architectures, which exploit quantum superposition for specific computational advantages. Currently, each represents evolutionary innovation \textit{within} the current lineage, not a new class of organism.

An era-level transition would require one or more of: (1) artificial general intelligence (AGI) operating reliably across all cognitive domains with human-level flexibility; (2) a fundamentally new learning paradigm replacing gradient-based backpropagation; (3) genuine recursive self-improvement, where AI systems modify their own architectures and training procedures to produce qualitatively more capable successors; or (4) autonomous AI-AI ecosystems with their own selection dynamics independent of human direction. None of these are imminent, but the accelerating pace of epoch-level transitions (the interval between GPT-3 in 2020 and o1 in 2024 was less than half the interval between AlexNet in 2012 and the transformer in 2017) suggests the next era-level transition may arrive sooner than historical intervals would predict. When it does, the Institutional Scaling Law will require re-derivation for the new substrate---but the framework's core insight, that institutional fitness is a function of environment-specific selection pressures and not merely of capability, will remain valid regardless of the underlying architecture.

\subsection{The Latent Capability Paradox}

A growing body of research and empirical evidence reveals that frontier models exhibit capabilities that were neither explicitly trained nor anticipated---persuasive strategies, implicit persona adaptation, sycophantic compliance patterns, and linguistic influence patterns that resist comprehensive characterization. Schaeffer et al.\ \cite{schaeffer2023} have challenged the ``emergent abilities'' narrative by showing that some apparent discontinuities in capability are artifacts of metric choice, but the consensus remains that sufficiently large models exhibit genuinely novel behaviors not predictable from smaller-scale training curves \cite{wei2022}. Concurrently, safety infrastructure has not scaled proportionally---safety teams at major labs have been reduced, researchers report commercial pressure overriding safety considerations, and red-teaming capacity lags model release cadence.

The convergence of expanding latent capabilities and eroding safety infrastructure is producing a trust deficit that itself becomes a \textit{selection pressure} in our evolutionary framework: nations seek auditable, controllable models (driving Sovereign AI adoption), enterprises seek reliable, bounded models (driving domain-specific deployment), and the market shifts toward smaller, verifiable, specialized alternatives (driving Symbiogenetic Scaling). The Institutional Fitness Manifold provides the formal structure for this dynamic: Capability-Trust Divergence (Theorem~1, Equation~\ref{eq:divergence}) predicts the sign-flip in fitness as trust erodes; Sequential Trust Degradation (Theorem~2, Equation~\ref{eq:trust_deg}) quantifies the compounding erosion across deployment contexts; and the Speciation Proposition (Equation~\ref{eq:speciation}) demonstrates that environmentally divergent trust requirements mathematically necessitate divergent model optima.

\section{Discussion}

The evidence assembled in Sections~3--6 strongly supports the characterization of AI development as a punctuated evolutionary process. Era-level transitions---the Dartmouth Conference (1956), the backpropagation revival (1986), AlexNet (2012), and the transformer (2017)---each produced discontinuous, irreversible shifts in the dominant computational paradigm, rendering the prior paradigm subordinate within years. Epoch-level transitions within the Generative AI era---GPT-3 (June 2020), ChatGPT (November 2022), and OpenAI o1 (September 2024)---reorganized the competitive landscape with accelerating frequency, confirming the punctuated equilibrium pattern at finer temporal scales. Independent empirical confirmation comes from the Epoch Capabilities Index (ECI) \cite{ho2025eci}: the statistically optimal breakpoint in frontier capability progression falls at April 2024, with the rate of improvement nearly doubling from $\sim$8 to $\sim$16 ECI points per year (Figure~\ref{fig:punctuated}). Notably, this breakpoint precedes the Epoch III--IV boundary we define at September 2024, suggesting that the capability acceleration served as a leading indicator of the qualitative shift into Symbiogenesis---the underlying selective pressures (increased RL investment, richer post-training regimes, maturing orchestration infrastructure) were already reshaping the fitness landscape before the defining architectural innovation of the new epoch arrived. This pattern---acceleration preceding punctuation---is itself consistent with the thermodynamic analogy: phase transitions in physical systems are preceded by precursor fluctuations that intensify as the critical threshold approaches.

The MIT NANDA \textit{State of AI in Business 2025} report \cite{challapally2025}---surveying thousands of enterprises across industries---found that 95\% of enterprise AI pilots produce zero measurable P\&L impact. The ``GenAI Divide'' between technical capability and business value reveals that institutional absorption of AI capabilities lags far behind the pace of technical innovation, confirming the Capability-Trust Divergence we formalize in Theorem~1. The report documents a ``shadow AI economy'' where 90\% of employees use personal AI tools while official enterprise initiatives stall---suggesting that adoption is demand-driven but deployment is institutionally bottlenecked by trust, governance, and integration challenges. The finding that external vendor partnerships succeed at twice the rate of internal AI builds (67\% vs.\ 33\%) suggests that co-evolutionary partnerships---where AI vendors adapt to institutional environments---may prove more adaptive than purely internal development, consistent with our Symbiogenetic Scaling prediction.

The Institutional Scaling Law (Equation~\ref{eq:inst_scaling}) provides the formal structure for the emerging bifurcation between capability and trust. At the optimal scale $N^*(\varepsilon)$, marginal capability gain exactly offsets marginal trust and cost penalties. Below $N^*$, the classical scaling paradigm holds: bigger is better. Above $N^*$, the Capability-Trust Divergence (Theorem~1) dominates: bigger is worse. The Symbiogenetic Scaling correction (Equation~\ref{eq:agent}) and the Convergence-Orchestration Threshold (Equation~\ref{eq:convergence}) formalize the alternative: orchestrated systems of domain-specific models---each operating below the trust-erosion threshold, tightly coupled to institutional tools and data, and coordinated through adaptive topology routing \cite{lu2026}---can exceed the institutional fitness of any individual frontier model.

The early 2026 developments documented in Section~\ref{sec:early2026} provide converging---though partial---evidence that the competitive landscape is shifting in the direction the framework's mathematical structure describes. The direction of investment at major labs has moved toward orchestration quality (Claude Code's agentic architecture, OpenClaw's skill-based agent composition) rather than exclusively toward increased model scale. The infrastructure for sub-frontier training has become radically more accessible (Karpathy's microgpt and nanochat projects). And composite systems are extending into physical deployment environments where the fitness weight vector $\w(\varepsilon)$ diverges sharply from text-based benchmarks. We emphasize that these observations are consistent with the framework's dynamics without constituting definitive empirical validation---the strongest test, whether orchestrated systems of small domain-specific models demonstrably outperform frontier generalists in documented institutional deployments, awaits systematic measurement of the kind outlined in our Remark following Definition~3.

The DeepSeek Moment (Section~4.5.1) deserves special emphasis as a case study. DeepSeek R1's release demonstrated frontier performance at roughly one-tenth to one-sixth of Western cost structures, and its open-source release under the MIT License forced a fundamental recalibration. By February 2026, Chinese open-source models on HuggingFace had surpassed Meta's Llama in cumulative downloads. DeepSeek simultaneously validated the open-source paradigm, undermined the assumption that export controls could contain Chinese AI capabilities, and triggered the largest single-day equity loss in stock market history---fulfilling the Phase Transition criterion (Equation~\ref{eq:phase}, $dH/dt \gg \lambda_{\text{crit}}$).

\textbf{Limitations.} Evolutionary metaphors, while powerful, are imperfect. Biological evolution is undirected; technological evolution involves deliberate design, capital allocation, and strategic intent. The ``fitness'' of an AI system is determined by economic and institutional selection pressures, not natural selection in the strict sense. Our taxonomy is necessarily retrospective, and era/epoch boundaries involve interpretive judgment. The early 2026 developments examined in Section~\ref{sec:early2026} are observational rather than experimental: we identify structural parallels between these developments and the framework's mathematical formalism, but we cannot rule out that the same observations would be equally consistent with alternative theoretical accounts. The strongest empirical test of the framework---controlled comparison of orchestrated domain-specific systems against frontier generalists across a representative sample of institutional deployment environments, with fitness measured along all four dimensions of Definition~1---remains a direction for future work. However, the formal mathematical framework (Definitions~1--3, Theorems~1--2, Propositions~1--2) operates independently of the metaphor, providing testable predictions about scaling behavior and speciation dynamics.

\section{Conclusion}

This paper has argued that AI development is best understood not as smooth, monotonic progress but as a punctuated process of stasis and sudden transition---and that the classical assumption of monotonic scaling is formally wrong for most institutional deployment environments. The Institutional Fitness Manifold extends Han et al.'s \cite{han2025} Sustainability Index from hardware-level to ecosystem-level analysis, introducing environment-dependent weights across four dimensions: capability, trust, affordability, and sovereign compliance. From this framework, the Capability-Trust Divergence theorem (Theorem~1) establishes that scaling up can reduce institutional fitness; the Speciation Proposition (Proposition~1) derives ecosystem fragmentation from environmental heterogeneity; and the Institutional Scaling Law (Proposition~2) identifies an environment-specific optimal scale $N^*(\varepsilon)$ that can fall well below the frontier. The Symbiogenetic Scaling correction (Equation~\ref{eq:agent}) carries the stronger result: the relevant unit of competition is not the individual model but the orchestrated system, and domain-specific models tightly coupled to institutional tools, data, and context can collectively exceed the fitness of any frontier generalist. Early 2026 developments in agentic orchestration, training democratization, and physical-world deployment (Section~\ref{sec:early2026}) exhibit these dynamics, though systematic empirical comparison of orchestrated domain-specific systems against frontier generalists in institutional deployments remains an open and critical direction for future work.

The classical scaling law says bigger is always better. The Institutional Scaling Law says \textit{better-adapted is always better}---and at the ecosystem level, adaptation increasingly means orchestrated specialization rather than undifferentiated scale. This distinction carries consequences beyond the technical. The hundreds of billions of dollars currently flowing into frontier model scale are premised on the monotonic assumption; the framework developed here suggests that a significant fraction of institutional value will instead accrue to systems engineered for specific deployment environments---systems that are smaller, more auditable, and sovereign by design. As Cellucci and Singh \cite{cellucci2025} observe, sovereignty will no longer be measured in land, assets, or GDP alone. It will be measured in who controls the intelligence that shapes the world. The nations, institutions, and organizations that recognize the non-monotonic logic of the Institutional Scaling Law---and invest accordingly in domain-adapted, orchestrated AI systems rather than pursuing scale for its own sake---will hold the decisive advantage in this new landscape.

\section*{Acknowledgements}

The author gratefully acknowledges Dr.\ Gunnar E.\ Carlsson (Stanford University), Dr.\ Anupam Chattopadhyay (Nanyang Technological University, Singapore), and Cardinal Peter Turkson (Chancellor of the Pontifical Academy of Sciences) for their valuable contributions and support.

\bibliographystyle{plain}

\end{document}